\documentclass{article} 
\usepackage{iclr2026_conference,times}

\usepackage{amsmath,amsfonts,bm}

\def\eqref#1{equation~\ref{#1}}

\def\1{\bm{1}}

\DeclareMathAlphabet{\mathsfit}{\encodingdefault}{\sfdefault}{m}{sl}
\SetMathAlphabet{\mathsfit}{bold}{\encodingdefault}{\sfdefault}{bx}{n}

\usepackage{hyperref}
\usepackage{url}

\usepackage[utf8]{inputenc}
\usepackage{booktabs}
\usepackage{makecell}
\usepackage{fontawesome5}
\usepackage{graphicx}

\usepackage{amsmath, amssymb}
\usepackage{booktabs}
\usepackage{multirow}
\usepackage{graphicx}
\usepackage{array}
\usepackage{tabularx}
\usepackage{pifont} 
\usepackage{float}
\usepackage{todonotes}
\usepackage{subcaption} 
\usepackage{array}
\usepackage{marvosym}

\title{PPE: Positional Preservation Embedding for Token Compression in Multimodal Large Language Models}

\author{Mouxiao Huang\thanks{Equal contribution.} ,
Borui Jiang\textsuperscript{\Letter}\footnotemark[1] ,
Dehua Zheng,
Hailin Hu\textsuperscript{\Letter},
Kai Han,
Xinghao Chen\textsuperscript{\Letter} \\
Huawei Technologies \\
\textsuperscript{\Letter}\texttt{\{jiangborui, hailin.hu, xinghao.chen\}@huawei.com}
}

\iclrfinalcopy 
\begin{document}

\maketitle

\begin{abstract}
Multimodal large language models (MLLMs) have achieved strong performance on vision-language tasks, yet often suffer from inefficiencies due to redundant visual tokens. Existing token merging methods reduce sequence length but frequently disrupt spatial layouts and temporal continuity by disregarding positional relationships. In this work, we propose a novel encoding operator dubbed as \textbf{P}ositional \textbf{P}reservation \textbf{E}mbedding (\textbf{PPE}), which has the main hallmark of preservation of spatiotemporal structure during visual token compression. PPE explicitly introduces the disentangled encoding of 3D positions in the token dimension, enabling each compressed token to encapsulate different positions from multiple original tokens. Furthermore, we show that PPE can effectively support cascade clustering --- a progressive token compression strategy that leads to better performance retention. PPE is a parameter-free and generic operator that can be seamlessly integrated into existing token merging methods without any adjustments. Applied to state-of-the-art token merging framework, PPE achieves consistent improvements of $2\%\sim5\%$ across multiple vision-language benchmarks, including MMBench (general vision understanding), TextVQA (layout understanding)  and VideoMME (temporal understanding). These results demonstrate that preserving positional cues is critical for efficient and effective MLLM reasoning. Our code is available at \url{https://github.com/MouxiaoHuang/PPE}.
\end{abstract}

\section{Introduction}
\label{intro}

\begin{figure*}[htbp]
  \centering
  \includegraphics[width=0.85\textwidth]{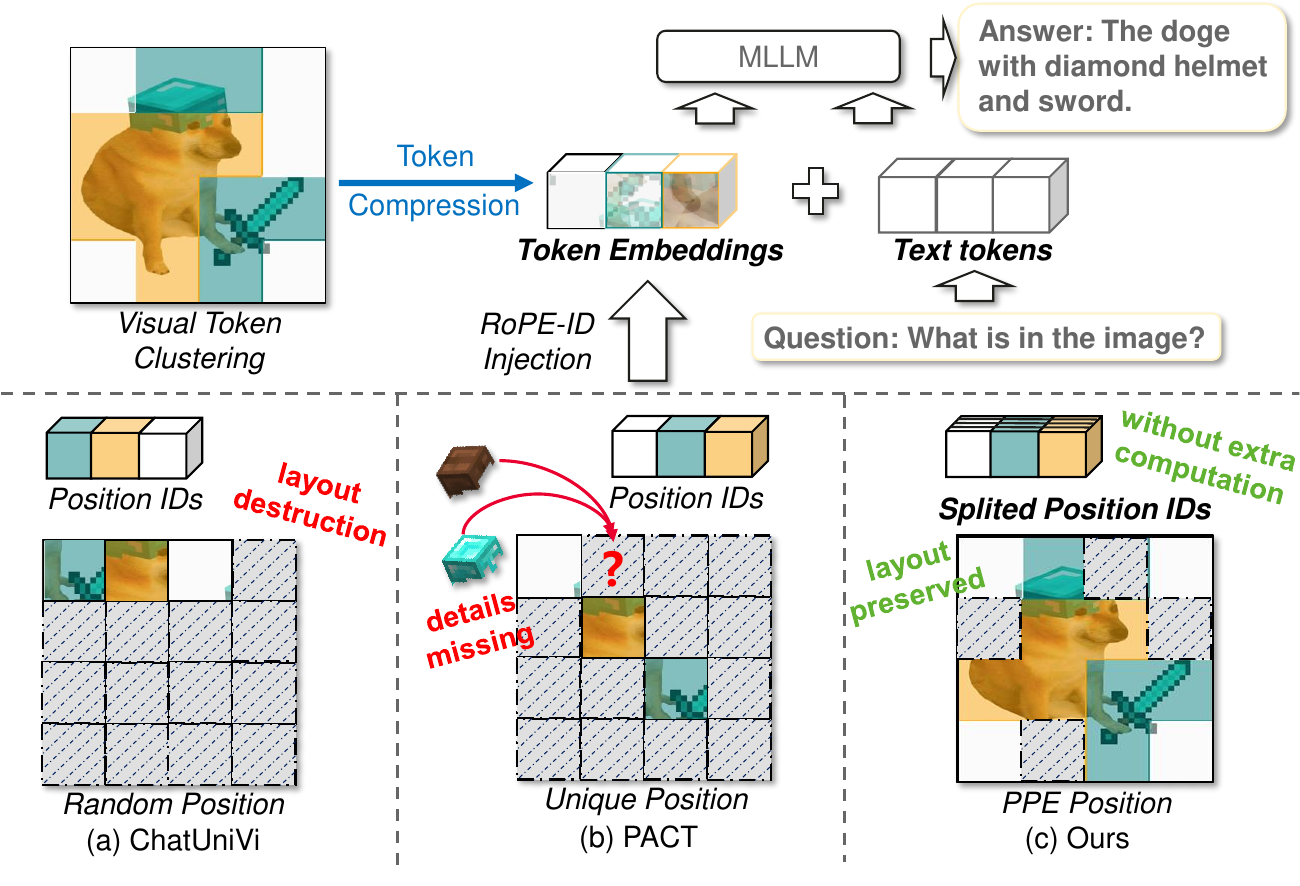}
  \caption{\textbf{Comparison between PPE and other token merging methods of processing positional IDs}. To simplify, the components such as the visual encoder are omitted. \textbf{(a)} ChatUniVi~\cite{jin2023chatunivi} mainly assigns randomize ID value to the clustered visual tokens. \textbf{(b)} PACT~\cite{dhouib2025pactpruningclusteringbasedtoken} retains the ID of the cluster center for the clustered visual tokens. \textbf{(c)} Proposed PPE splits the IDs of compressed token on different dimensions, so that each compressed token could contain \textbf{several} original position IDs. }
  \label{fig:idea}
  \vspace{-1.3em}
\end{figure*}

Multimodal Large Language Models (MLLMs) have recently achieved remarkable success across a range of vision-language understanding tasks~\cite{bai2025qwen25vltechnicalreport, lei2025scalabilitysimplicityempiricalanalysis, llava_onevision, Sa2VA, zhang2025pixelsailsingletransformerpixelgrounded}. A common paradigm involves encoding images or video frames into dense visual tokens, which are then fed into the language model for joint understanding. However, this dense representation is often highly redundant, leading to inefficiencies in computation and inference~\cite{song2024moviechatdensetokensparse}. To address this, recent works~\cite{dhouib2025pactpruningclusteringbasedtoken, jin2023chatunivi, ma2023imagesetpoints, zeng2022tokensequalhumancentricvisual, sparsevlm} have explored visual token compression, which merges similar tokens to reduce visual sequence length while preserving semantic information, thereby accelerating inference and lowering memory usage.

Despite the efficiency, existing compression methods often disrupt the spatial and temporal structure of visual inputs, limiting their applicability in layout-sensitive tasks such as counting, temporal grounding and sequential understanding. As shown in Figure~\ref{fig:idea}~(a), clustering-based Chat-UniVi~\cite{jin2023chatunivi} token compression may discard fine-grained spatial or temporal cues. Figure~\ref{fig:idea}~(b) illustrates the recent methods like PACT~\cite{dhouib2025pactpruningclusteringbasedtoken} which have attempted to preserve layouts during compression, but still remain constrained by the insufficient and imprecise positions.

In this work, we introduce \emph{Positional Preservation Embedding} (PPE), a novel positional encoding strategy which explicitly retains the different spatiotemporal layout into one compressed token. Our design is motivated by two key principles. First, we aim to preserve spatial and temporal positions during similar token merging. To this end, PPE firstly assigns each visual token a positional ID (\emph{e}.\emph{g}., 2D spatial for images and 3D spatiotemporal for videos). During compression, each merged token retains several positional IDs of their constituents. This ensures that most of the visual scene layouts are still accessible to MLLM at high compression rates, as shown in Figure~\ref{fig:idea}~(c).

Secondly, we support the widely adopted cascaded compression strategy, which performs token merging progressively across the Transformer~\cite{vaswani2023attentionneed} layers. This design is inspired by the observation that different layers capture increasingly abstract representations with higher similarity~\cite{bolya2023tokenmergingvitfaster, dhouib2025pactpruningclusteringbasedtoken, song2024moviechatdensetokensparse, sparsevlm}. Thus, merging tokens in a multi-stage manner enables higher compression ratios without collapsing the shallow semantics prematurely. Since PPE is decoupled from the merging algorithm and operates solely on position IDs, it naturally extends to multi-stage compression. This allows our method to preserve fine-grained spatiotemporal layouts across multiple compression stages, resulting in higher compression ratio, IDs retention~(\textit{Section~\ref{k_value}}) and improved performance~(\textit{Section~\ref{location}}).

To our best knowledge, PPE is the first work that explores an effective and lightweight positional preservation solution during visual token compression of MLLM. In contrast to existing positional embedding methods that only preserves one position for single token and lead to the loss of detailed layouts, PPE enables each single token to represent multiple positions, thus preserving the vision layout more completely. Moreover, PPE is parameter-free and can be used in a plug-and-play fashion for easily implanting into existing visual token compression methods without any additional computational costs. 

In the experiments, we apply PPE for tackling a variety of vision-language tasks, including general vision-language understanding on MMBench~\cite{liu2024mmbenchmultimodalmodelallaround}, text-based visual reasoning on TextVQA~\cite{singh2019vqamodelsread}, temporal understanding on VideoMME~\cite{fu2024videommefirstevercomprehensiveevaluation}, etc. The reported performances consistently outstrip previous compression methods~\cite{dhouib2025pactpruningclusteringbasedtoken,jin2023chatunivi} by significant margins after fine-tuning. We strongly believe that PPE can make inroads into domains of sparse token representing in MLLM where dense visual representation had previously reigned supreme.

The main contributions of this work are as follows:
\begin{itemize}
    \item We identify a critical limitation in existing visual token merging methods—namely, the neglect of spatial structure preservation and temporal coherence—which leads to distortion of intra-frame layouts and disruption of inter-frame temporal relations.
    \item We propose Positional Preservation Embedding (PPE), a novel, plug-and-play approach that explicitly preserves spatiotemporal integrity during token merging, effectively addressing both spatial and temporal challenges.
    \item We further show that PPE can be applied in cascade compression manner within multiple transformer layers of the MLLM, enabling substantial compression with minimal performance degradation.
    \item We conduct extensive experiments across serveal image and video benchmarks, demonstrating that PPE maintains accuracy while reducing the visual token count by 90\% and outperforms other visual token compression methods at comparable reduction ratios.
    \vspace{-0.6em}
\end{itemize}

\section{Related Work}
\label{related_work}

\subsection{Multimodal Large Language Models}

MLLMs extend traditional LLMs by incorporating visual inputs, enabling unified vision-language understanding and generation. Recent models such as Flamingo~\cite{alayrac2022flamingovisuallanguagemodel}, the LLaVA series~\cite{llava_onevision, liu2023visualinstructiontuning, zhang2024video-llava-video}, and the Qwen-VL series~\cite{bai2023qwenvlversatilevisionlanguagemodel, bai2025qwen25vltechnicalreport, wang2024qwen2vlenhancingvisionlanguagemodels} achieve strong performance across captioning, VQA, and instruction following. These models typically align vision and language via cross-attention, projection modules, or lightweight multimodal transformers. However, most rely on dense visual tokens, posing efficiency and scalability challenges for high-resolution or long-form visual inputs. Extensions such as Video-LLaVA~\cite{lin2024videollavalearningunitedvisual}, VideoChat~\cite{li2024videochatchatcentricvideounderstanding}, and Video-LLaMA~\cite{zhang2023videollamainstructiontunedaudiovisuallanguage} mitigate this via frame sampling or sparse memory. In contrast, our method introduces a Positional Preservation Embedding that enables substantial token reduction while maintaining spatiotemporal coherence, offering a scalable alternative for multimodal reasoning.

\subsection{Visual Token Merging}
To improve the efficiency of MLLMs, recent works have explored visual token reduction strategies~\cite{DBLP:journals/corr/abs-2505-18227} such as clustering and merging~\cite{bolya2023tokenmergingvitfaster, xu2022groupvitsemanticsegmentationemerges, ma2023imagesetpoints, zeng2022tokensequalhumancentricvisual, dhouib2025pactpruningclusteringbasedtoken, song2024moviechatdensetokensparse, jin2023chatunivi, visionzip}. These methods significantly reduce the sequence length of image or video inputs, enabling faster inference and reduced memory consumption. Techniques include token matching~\cite{bolya2023tokenmergingvitfaster}, hierarchical clustering~\cite{ma2023imagesetpoints, zeng2022tokensequalhumancentricvisual}, group-based representation~\cite{xu2022groupvitsemanticsegmentationemerges}, and pruning-clustering hybrids~\cite{dhouib2025pactpruningclusteringbasedtoken}, with extensions to long-video understanding via sparse memory representations~\cite{song2024moviechatdensetokensparse} and unified token compression across modalities~\cite{jin2023chatunivi}. Despite their effectiveness, these methods typically discard original positional information, which can disrupt spatial layouts and temporal continuity—limiting performance on fine-grained visual reasoning tasks. In contrast, our approach explicitly preserves spatiotemporal position cues throughout compression, maintaining structural fidelity while achieving substantial token reduction.

\subsection{Positional Encoding in MLLMs}
Positional encoding~\cite{vaswani2023attentionneed} is essential in MLLMs to maintain spatial and temporal relationships across vision and language tasks. The Rotary Position Embedding (RoPE)~\cite{su2023roformerenhancedtransformerrotary} is a widely adopted method, which captures relative token positions in sequences using a rotational mechanism. This has been extended to 2D RoPE~\cite{heo2024rotarypositionembeddingvision}, which is well-suited for image tasks, enabling models to understand spatial locality in vision transformers (ViTs). For video and temporal tasks, the Qwen series introduced 3D MRoPE~\cite{bai2023qwenvlversatilevisionlanguagemodel, wang2024qwen2vlenhancingvisionlanguagemodels}, which integrates rotary encoding across both spatial and temporal dimensions, preserving continuity across frames. However, these methods often face challenges in visual token compression, as reducing token counts can lead to the loss of crucial positional information. In contrast, we propose \textbf{PPE}: \textbf{P}ositional \textbf{P}reservation \textbf{E}mbedding, a novel approach that maintains spatiotemporal positional cues during token reduction, preserving structural integrity while enabling aggressive compression. This method, to the best of our knowledge, is a new exploration in the field.

\begin{figure*}[htbp]
  \centering
  \includegraphics[width=0.99\textwidth]{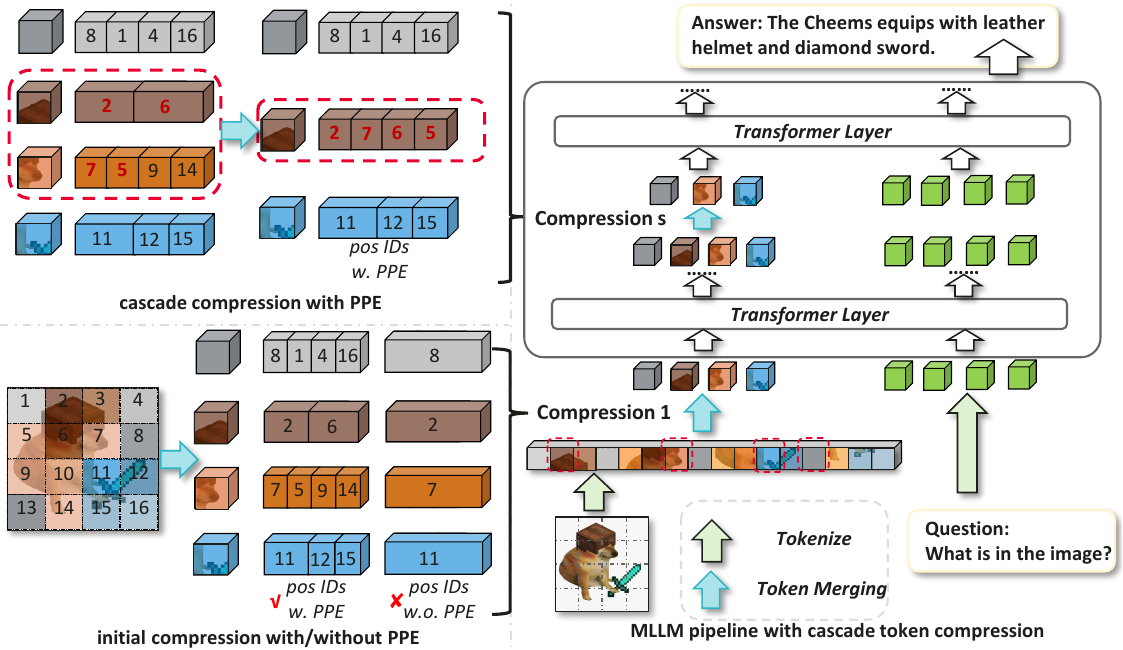}
  \caption{\textbf{The overview pipeline of the proposed PPE with cascade compression.} \textbf{Left:} Main idea of Positional Preservation Embedding (PPE) integrated in token compression. For each RoPE ID in compressed token embedding, PPE splits the dimension into chunks to prefill multiple position IDs. The IDs of tokens with high importance scores are reserved preferentially. \textbf{Right:} The MLLM pipeline integrating PPE and cascade compression. Token compression is applied in multiple layers, each with PPE. See main text for more explanation. }
  \label{fig:fig2}
\end{figure*}

\section{Methodology}
\label{method} 

In this section, we first recap the preliminaries of PPE, and then analyze why previous methods fail to hold the positional information. Afterward, we introduce our proposed PPE and cascade compression, as illustrated in Figure~\ref{fig:fig2}.

\subsection{Preliminary}

\textbf{Rotary Position Embeddings in MLLM. } RoPE~\cite{su2023roformerenhancedtransformerrotary} is a positional encoding technique designed to enhance the Transformer~\cite{vaswani2023attentionneed} architecture by integrating relative positional information directly into the self-attention mechanism. Unlike traditional absolute positional encodings, RoPE applies a rotation to the query and key vectors in multi-head attention, enabling the model to capture relative positions effectively. Assumed that token vector $\mathbf{z} \in \mathbb{N}^D$ in multi-head attention, the rotation operation can be represented as:
\begin{equation}
    \text{RoPE}(z_d, m) = e^{i m \theta_d} z_d, ~d = 1 ... D,
\label{equ1}
\end{equation}
where $m$ indicates the position ID of $z$. For capturing spatiotemporal layouts, Qwen2.5-VL~\cite{bai2025qwen25vltechnicalreport} introduces M-RoPE, a structured embedding scheme that partitions the embedding dimension, each encoding positional information along different visual axis—such as temporal order, image height, and width. Formally, the original RoPE rotation operation is modified to:
\begin{equation}
    \text{M-RoPE}(z_d, m_d) = e^{i m_d \theta_d} z_d, ~d = 1 ... D,
\label{equ2}
\end{equation}
where $\mathbf{m} \in \mathbb{Z}^D$ is pre-filled 2D or 3D visual position IDs. Assumed that the 3D visual token is at position $(t, h, w)$, the $\{m_d\}$ could be indicated as:
\begin{equation}
m_d = 
\begin{cases}
t, & d = \;1 \;...\; D_1, \\
h, & d = \;D_1+1  \;...\; D_1+D_2, \\
w, & d = \;D_1+D_2+1  \;...\; D_1+D_2+D_3,
\end{cases}
\end{equation}
where $D_1, D_2, D_3$ are human-crafted integer to control the size of mrope sections~\cite{bai2025qwen25vltechnicalreport} satisfying that $D_1+D_2+D_3=D$.

\textbf{Visual Token Compression.} ChatUniVi~\cite{jin2023chatunivi} adopts a lightweight, parameter-free clustering algorithm for token compression to mitigate the computational overhead of long visual token sequences. The key idea is to merge tokens with DPC-KNN~\cite{dpcknn} based on token similarity. Assumed that $N$ visual token embeddings $\{\mathbf{z}_i \in \mathbb{R}^D\}_{i=1}^N$ are clustered to $M$ groups, $\{\mathbf{z}_i\}_{i \in C_j}$ indicates the token embeddings in group $j$. The compressed token embeddings $\mathbf{z}_j'$ is defined as:
\begin{equation}
    \mathbf{z}_j' = \frac{1}{|C_j|} \sum_{i \in C_j} \mathbf{z}_i, ~j=1...M.
\end{equation}

\subsection{Positional Preservation Embedding} 
\label{sec:positionid} 

While similarity-based compression effectively reduces sequence length, it disrupts fine-grained spatiotemporal layouts. To mitigate this, we propose \textit{Positional Preservation Embedding} (PPE), which retains multiple positional cues per merged token to preserve structural information.

PPE builds on the principle behind RoPE~\cite{su2023roformerenhancedtransformerrotary}, in which the rotation of position embeddings is independent at the dimension. As a result, the $m$ in Equation~\ref{equ1} can be totally different on the dimension $D$ to represent different positions at the same time. For instance, M-RoPE~\cite{wang2024qwen2vlenhancingvisionlanguagemodels} partitions embedding dimensions into several groups to store spatiotemporal positions, formally in Equation~\ref{equ2}. Inspired by this, PPE merges different positions to one merged token by splitting into more groups. Formally, the merged PPE ID $\hat{\mathbf{m}}$ could be indicated as:
\begin{equation}
    \hat{m}_d = m_{k,d},\; d = (k-1)\frac{D}{K}+1 \;...\; k\frac{D}{K},
\end{equation}
where $K$ is a fixed hyper-parameter to represent the maximum capacity of PPE, and $\{\mathbf{m}_k\}^K$ is the set of different position IDs in 1D RoPE manner before merging. Note that $K$ is always divisible by dimension $D$. In M-RoPE manner, $K$ is set to the greatest common divisor of the mrope sections, guaranteed that each dimension is evenly cut into $K$ groups:
\begin{equation}
    \hat{m}_d^{3D} = m_{k,d}^{3D}, \; d = 
\begin{cases}
(k-1)\frac{D_1}{K}+1 & ... \;\; k\frac{D_1}{K}, \\
(k-1)\frac{D_2}{K}+D_1+1& ...\;\; k\frac{D_2}{K}+D_1,\\
(k-1)\frac{D_3}{K}+D_1 + D_2 + 1& ...\;\; k\frac{D_3}{K}+D_1 + D_2.\\
\end{cases}
\end{equation}

The key insight shared of PPE is that similar token embeddings can share their feature embeddings during token merging. Rather than assigning a single position ID to a merged token, we extend this idea to a multi-position formulation that better reflects the internal diversity of a token cluster - containing more positional information of original input visual tokens. Specifically, consider a cluster group $C_j$ containing token embeddings $\{\mathbf{z}_i\}_{i\in C_j}$ with corresponding position IDs $\{\mathbf{m}_i$. Rather than choosing one representative ID, we select the top-$K$ IDs per cluster which are scored by the distance from the cluster center. Note that the score is higher if the token is closer to the cluster center~\cite{jin2023chatunivi}. If $|C_j| < K$, high-weight tokens are repeated to fill the slots. Denote the merged IDs of PPE as $\hat{\mathbf{m}}$, the PPE rotation of vector $\mathbf{z}$ is simply follow the RoPE which could be formulated as:
\begin{equation}
    \mathrm{PPE}(z_d, \hat{m}_d) \;=\;\,e^{i\,\hat{m}_d\theta_d}\,z_d, ~d = 1 ... D,
\end{equation}

\subsection{Cascade Compression with PPE}

In this section, we further investigate the effectiveness of cascade compression in conjunction with our proposed Positional Preservation Embedding (PPE) strategy. Cascade compression is a widely adopted technique in token compression~\cite{bolya2023tokenmergingvitfaster, dhouib2025pactpruningclusteringbasedtoken, song2024moviechatdensetokensparse, sparsevlm}, motivated by the observation that deeper Transformer layers exhibit greater representational redundancy~\cite{DBLP:conf/icml/DongCL21}, whereas earlier layers encode critical low-level semantics that are less amenable to aggressive compression.

To this end, we implement a cascaded PPE-based compression pipeline built upon ChatUniVi~\cite{jin2023chatunivi}, as illustrated in Figure~\ref{fig:fig2}. In this design, visual token clustering is applied not only prior to feeding tokens into the LLM, but also within selected LLM layers. Leveraging PPE, we preserve original positional information, enabling efficient computation of new cluster center positions after each merge.

For PPE ID assignment, we follow the standard top-K selection strategy described in \textit{Section~\ref{sec:positionid}}, choosing token IDs with minimal distance to the cluster center. As shown in Figure~\ref{fig:fig2}, during repeated merges, the number of preserved position IDs reduces gradually (e.g., retaining only four IDs when two previously merged tokens are merged again).

Experimental results demonstrate that this cascaded PPE design maintains fine-grained spatial structure across compression stages and achieves higher compression ratio without performance loss. Moreover, this design significantly improves both ID retention (\textit{Section~\ref{k_value}}) and downstream task performance (\textit{Section~\ref{location}}). These findings confirm the strong compatibility of PPE with standard cascaded compression frameworks.

\section{Experiment}
\label{exp}

\subsection{Experimental Setup}  
\textbf{Datasets.} To demonstrate the effectiveness of PPE, we construct our training datasets for supervised fine-tuning (SFT) referring to the public projects: LLaVA-Video-178k~\cite{zhang2024video-llava-video} and LLaVA-OneVision~\cite{llava_onevision}. Limited by the computational costs, we adopted even down-sampling on these datasets. To highlight the improvements of the model on image and video benchmarks, we further set up two different settings: one for images and another for videos. Specifically, we utilized about 120K video samples to fine-tune the models for multimodal video benchmarks, while 300K image samples to the models for the multimodal image benchmarks.

\textbf{Evaluation Benchmarks.} We use VideoMME~\cite{fu2024videommefirstevercomprehensiveevaluation} as our primary benchmark for multimodal video analysis. To assess generalization, we also report results on NeXT-QA (multi-choice and open-ended)~\cite{xiao2021nextqanextphasequestionansweringexplaining}, SEED‑Bench‑Video~\cite{li2023seedbenchbenchmarkingmultimodalllms}, and MVBench~\cite{li2024mvbenchcomprehensivemultimodalvideo}. For image understanding, we evaluate on MMBench (CN/EN)~\cite{liu2024mmbenchmultimodalmodelallaround} and SQA~\cite{iyyer2017search} to assess general visual comprehension. To further examine text-rich layout and document understanding, we include TextVQA~\cite{singh2019vqamodelsread}, DocVQA~\cite{docvqa}, OCRBench~\cite{ocrbench}, and ChartQA~\cite{chartqa}. All reported metrics follow the evaluation criteria of their respective benchmarks, and unless otherwise noted, higher values indicate better performance.

\textbf{Implementation Details.} We conduct our default experimental settings by using the Qwen2.5‑VL‑ 3B‑Instruct model~\cite{bai2025qwen25vltechnicalreport}. All models are fine-tuned in a fully supervised manner on the down-sampled datasets with all parameters unfrozen. The training setting is mainly referring to the public project~\cite{Qwen2-VL-Finetuning}, in which the gradient accumulation step is $4$, the learning rate is $1\mathrm{e}{-5}$ along with the warm-up ratio of $0.03$, while the learning rate of visual encoder and patch merger is $2\mathrm{e}{-6}$ and $1\mathrm{e}{-5}$, respectively. We train for only $1$ epoch. To maintain the optimal performance and computational costs for both training and evaluation, we mainly follow the official input size configurations. Specifically, we set \texttt{image\_min\_pixels} = $512 \times 28 \times 28$ and \texttt{image\_max\_pixels} = $1280 \times 28 \times 28$ for image inputs, and \texttt{video\_min\_pixels} = $128 \times 28 \times 28$ and \texttt{video\_max\_pixels} = $768 \times 28 \times 28$ for video inputs. Additionally, the maximum number of video frames is set to $64$ due to the limited memory usage. The 3D M-RoPE section is set to $[16, 24, 24]$, while $[32, 32]$ for the 2D manner. The number of preserved token IDs is set to $K=8$ for 3D and $K=32$ for 2D, which corresponds to the greatest common divisor (GCD) of M-RoPE sections.

\textbf{Token compression settings.} The proposed PPE can be integrated into most existing compression methods. For our experiments, we adopt the SOTA clustering-based framework Chat-UniVi~\cite{jin2023chatunivi} and follow its default training strategy, with slight modifications to adapt to the Qwen2.5-VL model. In Chat-UniVi, the number of clustered tokens is controlled by a predefined clustering ratio due to the native resolution technique. Specifically, the spatial clustering ratio is set to $0.45$, while the temporal clustering ratio is $0.0625$, consistent with the original paper. For fair comparison, token clustering is applied at the interface between the vision encoder and the LLM by default.

\subsection{Main Results}

We evaluate the performance of our proposed method on a range of image and video benchmarks. The results show that PPE consistently outperforms previous approaches, demonstrating its effectiveness in both image and video understanding.

\begin{table}[t]
\centering
\caption{The overall performance comparison across different benchmarks. The \textbf{Dense} model is a simply Qwen2.5-VL-3B-Instruct model fine-tuned on our SFT datasets. The subscript S/ST indicates spatial-only and spatiotemporal compression, respectively. \textbf{Bold} denotes the best performance among compressed variants, and \underline{underline} denotes the best overall including the Dense baseline. The image average score is calculated without OCRBench.}
\small
\begin{tabular}{@{}cccc|cc@{}}
\toprule
\textbf{Benchmarks} & \textbf{Dense} & \textbf{+Chat-UniVi$_{\text{S}}$} & \textbf{+PPE$_{\text{S}}$ (Ours)} & \textbf{+Chat-UniVi$_{\text{ST}}$} & \textbf{+PPE$_{\text{ST}}$ (Ours)} \\
\cmidrule(lr){1-1} \cmidrule(lr){2-6}
\textbf{MMBench} (EN) & \underline{85.89} & \textbf{84.92} & 84.73~(-0.19) & - & - \\
\textbf{MMBench} (CN) & \underline{86.07} & 83.71 & \textbf{84.87}~(+1.16) & - & - \\
\textbf{SQA} & 76.90 & 77.30 & \textbf{\underline{77.88}}~(+0.58) & - & - \\
\textbf{TextVQA} & \underline{79.50} & 57.66 & \textbf{77.14}~(+19.48) & - & - \\
\textbf{DocVQA} & \underline{89.44} & 52.48 & \textbf{76.79}~(+24.31) & - & - \\
\textbf{OCRBench} & \underline{761} & 535 & \textbf{691}~(+156) & - & - \\
\textbf{ChartQA} & \underline{79.96} & 49.60 & \textbf{74.52}~(+24.92) & - & - \\
\cmidrule(lr){1-1} \cmidrule(lr){2-6}
\textit{\textbf{Image Average}} & \underline{\textit{82.96}} & \textit{67.61} & \textbf{\textit{79.32}}~(+11.71) & - & - \\
\midrule
\textbf{VideoMME} (w/o subs) & 57.81 & 57.22 & \textbf{\underline{58.70}}~(+1.48) & 56.07 & \textbf{57.41}~(+1.34) \\
\textbf{VideoMME} (w subs) & 57.96 & 57.22 & \textbf{\underline{59.07}}~(+1.85) & 56.15 & \textbf{57.78}~(+1.63) \\
\textbf{NeXT-QA} (MC) & 78.20 & 77.63 & \textbf{\underline{78.42}}~(+0.42) & 77.59 & \textbf{77.99}~(+0.4) \\
\textbf{NeXT-QA} (OE) & 31.65 & 25.37 & \textbf{\underline{32.61}}~(+7.24) & 26.55 & \textbf{31.95}~(+5.4) \\
\textbf{SEED-Bench-Video} & \underline{57.60} & \textbf{56.08} & 55.98~(-0.10) & 53.47 & \textbf{54.19}~(+0.72) \\
\textbf{MVBench} & \underline{67.90} & 66.90 & \textbf{67.38}~(+0.48) & 64.38 & \textbf{66.42}~(+2.04) \\
\cmidrule(lr){1-1} \cmidrule(lr){2-6}
\textit{\textbf{Video Average}} & \textit{58.52} & \textit{56.74} & \textbf{\textit{\underline{58.69}}}~(+1.95) & \textit{55.70} & \textbf{\textit{57.62}}~(+1.92) \\
\midrule
\textbf{Reduction Ratio} & 0\% & 55\% & 55\% & 94\% & 94\% \\
\bottomrule
\end{tabular}
\label{tab:main_table}  
\vspace{-1.3em}
\end{table}

\textbf{Image Tasks.} Table~\ref{tab:main_table} illustrates the strong performance of PPE on image understanding benchmarks. Despite reducing visual tokens by 55\%, PPE achieves competitive or superior performance across all tasks. It significantly outperforms Chat-UniVi in overall accuracy (\textbf{79.32\%} vs. \textbf{67.61\%}) and on TextVQA (\textbf{77.14\%} vs. \textbf{57.66\%}), while maintaining comparable results on other tasks. Compared to the full-token Dense baseline model, PPE shows only a minor acceptable performance drop in overall accuracy (\textbf{82.96\%} vs. \textbf{79.32\%}), despite using less than half the visual tokens.

\textbf{Video Tasks.} As shown in Table~\ref{tab:main_table}, under a 55\% token reduction with spatial-only compression, PPE consistently outperforms Chat-UniVi across most tasks and in overall accuracy (\textbf{58.69\%} vs. \textbf{56.74\%}), even exceeding the Dense baseline. With a more aggressive 94\% reduction using spatiotemporal compression, PPE again leads in overall score (\textbf{57.32\%} vs. \textbf{55.7\%}), demonstrating strong robustness under heavy compression. Notably, on challenging benchmarks like VideoMME and NeXT-QA (OE), PPE shows clear improvements over Chat-UniVi.

\textbf{Summary.} In summary, PPE achieves substantial token reduction—55\% for spatial-only compression and 94\% for spatiotemporal compression—while preserving, or in some cases even enhancing, downstream task performance. This demonstrates that preserving visual token IDs enables more accurate reconstruction of the spatial layout and temporal order, even under aggressive compression, thereby improving reasoning capabilities, as illustrated in Fig.~\ref{fig:idea}.

\textbf{Comparison with M-RoPE.} As reported in Table~\ref{tab:main_table}, Dense model adopts M-RoPE by default, and thus Chat-UniVi directly inherits this design. In contrast, our PPE strategy compresses visual tokens while still retaining the positional information of merged tokens within a single representation. This design leads to consistently stronger results under identical reduction ratios, highlighting that PPE is more effective than M-RoPE in preserving positional cues during token compression.

\begin{table}[t]
\centering
\caption{Comparison of MLLMs with intact tokens and compressed tokens.}
\resizebox{\columnwidth}{!}{%
\begin{tabular}{ccccccccc}
\toprule
\textbf{Model} & \begin{tabular}[c]{@{}c@{}}\textbf{VideoMME}\\ (w/o subs)\end{tabular} & \begin{tabular}[c]{@{}c@{}}\textbf{VideoMME}\\ (w subs)\end{tabular} & \textbf{MVBench} & \begin{tabular}[c]{@{}c@{}}\textbf{MMBench}\\ (EN)\end{tabular} & \begin{tabular}[c]{@{}c@{}}\textbf{MMBench}\\ (CN)\end{tabular} & \textbf{TextVQA} & \begin{tabular}[c]{@{}c@{}}\textbf{Reduction}\\ \textbf{Ratio}\end{tabular} \\

\cmidrule(lr){1-1}\cmidrule(lr){2-4}\cmidrule(lr){5-7}\cmidrule(lr){8-8}
LLVA-OneVision-0.5B & 44.00 & 43.50 & 45.50 & 52.10 & - & - & 0\% \\
InternVL2.5-4B & 62.30 & 63.60 & 71.60 & 81.10 & 79.30 & 76.80 & 0\% \\
Qwen2.5-VL-3B & 61.50 & 67.60 & 67.00 & 79.10 & 78.10 & 79.30 & 0\% \\

\midrule
PACT-7B & 57.60 & - & - & 80.30 & - & 75.00 & 67\% \\
SparseVLM-7B & - & - & - & 64.10 & - & 57.80 & 66\% \\
\textbf{PPE-3B (Ours)} & \textbf{58.70} & \textbf{59.07} & \textbf{67.38} & \textbf{84.78} & \textbf{84.85} & \textbf{77.08} & 55\% \\
\textbf{PPE*-3B (Ours)} & 58.48 & 58.52 & 67.35 & - & - & - & 90\% \\

\bottomrule 
\end{tabular}%
}
\label{tab:comparison}
\vspace{-0.6em}
\end{table}

\begin{table}[t]
\centering
\caption{Performance of cascade compression, PPE integrated \textit{Before} and/or \textit{Within}-LLM.}
\resizebox{\columnwidth}{!}{%
\begin{tabular}{cccccccccc}
\toprule
\textbf{\textit{Before}} & \textbf{\textit{Within}} & \begin{tabular}[c]{@{}c@{}}\textbf{VideoMME}\\ (w/o subs)\end{tabular} & \begin{tabular}[c]{@{}c@{}}\textbf{VideoMME}\\ (w subs)\end{tabular} & \begin{tabular}[c]{@{}c@{}}\textbf{NeXT-QA}\\ (MC)\end{tabular} & \begin{tabular}[c]{@{}c@{}}\textbf{NeXT-QA}\\ (OE)\end{tabular} & \begin{tabular}[c]{@{}c@{}}\textbf{SEED-Bench}\\ \textbf{-Video}\end{tabular} & \textbf{MVBench} & \textbf{\textit{Average}} & \begin{tabular}[c]{@{}c@{}}\textbf{Reduction}\\ \textbf{Ratio}\end{tabular} \\
\cmidrule(lr){1-2}\cmidrule(lr){3-8}\cmidrule(lr){9-9}\cmidrule(lr){10-10}
\ding{55} & \ding{55} & 56.41 & 56.49 & 78.07 & 32.99 & 55.12 & 68.25 & \textit{57.91} & 0\% \\
\ding{51} & \ding{55} & 58.70 & 59.07 & 78.42 & 32.61 & 55.98 & 67.38 & \textit{58.69} & 55\% \\
\cmidrule(lr){1-2}\cmidrule(lr){3-8}\cmidrule(lr){9-9}\cmidrule(lr){10-10}
\ding{51} & \ding{55} & 57.41 & 57.70 & 77.98 & 31.06 & 55.36 & 66.65 & \textit{57.69} & 90\% \\
\ding{55} & \ding{51} & 58.48 & 58.52 & 78.20 & 32.20 & 56.11 & 67.35 & \textit{58.48} & 90\% \\
\bottomrule
\end{tabular}%
}
\label{tab:location}
\end{table}

\begin{table}[t]
\centering
\caption{Ablation of PPE integration with clustering-based compression and inference efficiency.}
\resizebox{\columnwidth}{!}{%
\begin{tabular}{cccccccc}
\toprule
\textbf{Method} 
& \begin{tabular}[c]{@{}c@{}}\textbf{MMBench}\\ (EN)\end{tabular} 
& \begin{tabular}[c]{@{}c@{}}\textbf{MMBench}\\ (CN)\end{tabular} 
& \textbf{TextVQA} 
& \begin{tabular}[c]{@{}c@{}}\textbf{LLM Generation (s)}\\ \textbf{Time (s) ↓}\end{tabular} 
& \begin{tabular}[c]{@{}c@{}}\textbf{Peak Memory}\\ \textbf{Usage (GB) blue}{↓}\end{tabular} 
& \begin{tabular}[c]{@{}c@{}}\textbf{Reduction}\\ \textbf{Ratio}\end{tabular} \\

\cmidrule(lr){1-1}\cmidrule(lr){2-5}\cmidrule(lr){6-7}
PACT & 74.14 & 74.17 & 73.73 & 0.08 & 15.82 & 89\% \\
PACT + PPE & \textbf{74.48} & \textbf{75.00} & \textbf{73.87} & 0.09 & 15.82 & 89\% \\
\midrule
ToMe & 74.31 & 73.63 & 74.94 & 0.90 & 15.81 & 57\% \\
ToMe + PPE & \textbf{74.57} & \textbf{74.74} & \textbf{76.16} & 0.91 & 15.81 & 57\% \\

\bottomrule 
\end{tabular}%
}
\label{tab:other}
\vspace{-0.6em}
\end{table}

\textbf{Comparison of MLLMs with Intact and Compressed Tokens.} We compare our PPE with representative dense MLLMs, including LLaVA-OneVision~\cite{llava_onevision}, InternVL2.5~\cite{internvl25}, and Qwen2.5-VL~\cite{bai2025qwen25vltechnicalreport}, as well as token compression approaches such as PACT~\cite{dhouib2025pactpruningclusteringbasedtoken} and SparseVLM~\cite{sparsevlm}, as shown in Table~\ref{tab:comparison}. All baseline results are reported by their respective original papers. Note that Chat-UniVi~\cite{jin2023chatunivi} is excluded due to missing benchmark results. Despite using only 3B parameters, PPE achieves competitive performance compared to both dense and compressed models, with a 55\% token reduction ratio. PPE* refers to a cascade compression variant applied \textit{within} the LLM (details in \textit{Section~\ref{location}}), which further improves the reduction ratio to 90\% while maintaining comparable performance.

\subsection{Cascade Compression}  
\label{location}

In addition to applying our PPE compression \textit{before} the LLM, we further investigate its integration \textit{within} the LLM to enable multi-layer cascade token compression. Specifically, we conduct a layer-wise insertion study using the Qwen2.5-VL-3B-Instruct model, which contains 36 transformer layers. Specifically, we insert PPE-based clustering modules at layers 11, 23, and 35, while keeping all other configurations unchanged.

As shown in Table~\ref{tab:location}, applying PPE \textit{within} the LLM using a per-layer clustering ratio of 0.45 achieves a 90\% token reduction while maintaining performance comparable to the 55\%-reduction case. Under the same 90\% reduction budget, the \textit{within}-LLM configuration outperforms applying PPE only \textit{before} the LLM, demonstrating that cascade compression across multiple stages enables higher compression ratios without prematurely collapsing shallow semantics.

\subsection{Ablation Study}

In this section, we study the insight of PPE, its compatibility with other clustering-based compression methods, inference efficiency, and the effect of retaining different numbers of token ID positions to verify its ability to preserve positional information. Further studies on different reduction ratios~\textit{\ref{append_ratio}}, aggregation stages~\textit{\ref{append_stages}}, model size generalization~\textit{\ref{append_model_size}}, backbone generalization~\textit{\ref{append_backbone}}, task generalization~\textit{\ref{append_task_generalization}}, etc, are provided in the \textit{Appendix~\ref{append}}.

\subsubsection{Insight into why PPE better preserves token semantics}
\label{sec_ppe_insight}
The key insight of PPE lies in its ability to simultaneously encode multiple positions within a single token. Specifically, unlike traditional RoPE where multiple frequencies are rigidly bound to a single positional index, PPE introduces flexibility by allowing different frequencies to bind to different positional indices—as long as the corresponding tokens are considered similar in the clustering. As a result, during attention computation, each token can attend to multiple combinations of frequencies and positional indices. This enables each token pair to perceive a richer set of relative positional relationships, thereby preserving the global positional layout despite compression.

\begin{figure*}[htbp]
  \vspace{-0.6em}
  \centering
  \begin{subfigure}{0.25\textwidth}
    \centering
    \includegraphics[width=\linewidth]{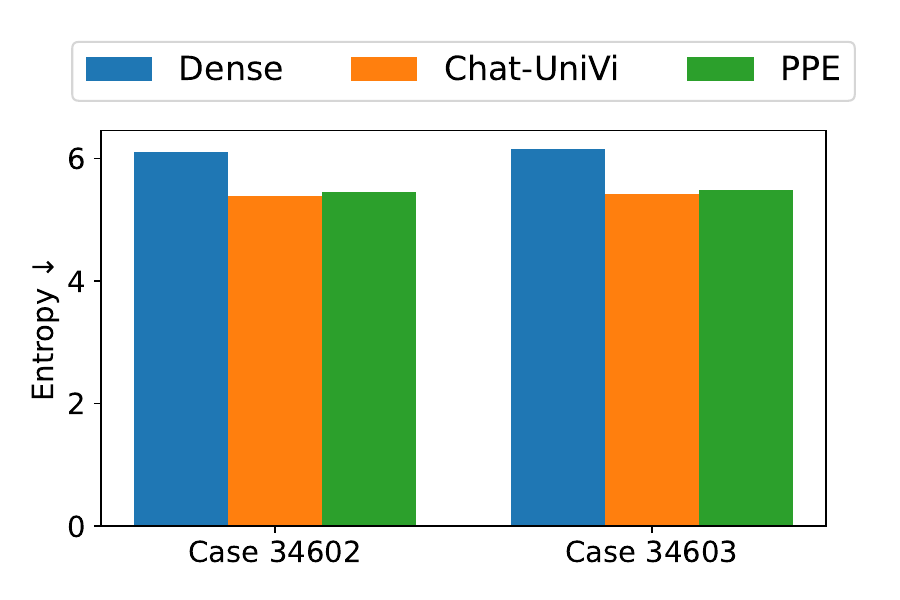}
    \caption{Entropy comparison ↓}
  \end{subfigure}
  \hfill
  \begin{subfigure}{0.25\textwidth}
    \centering
    \includegraphics[width=\linewidth]{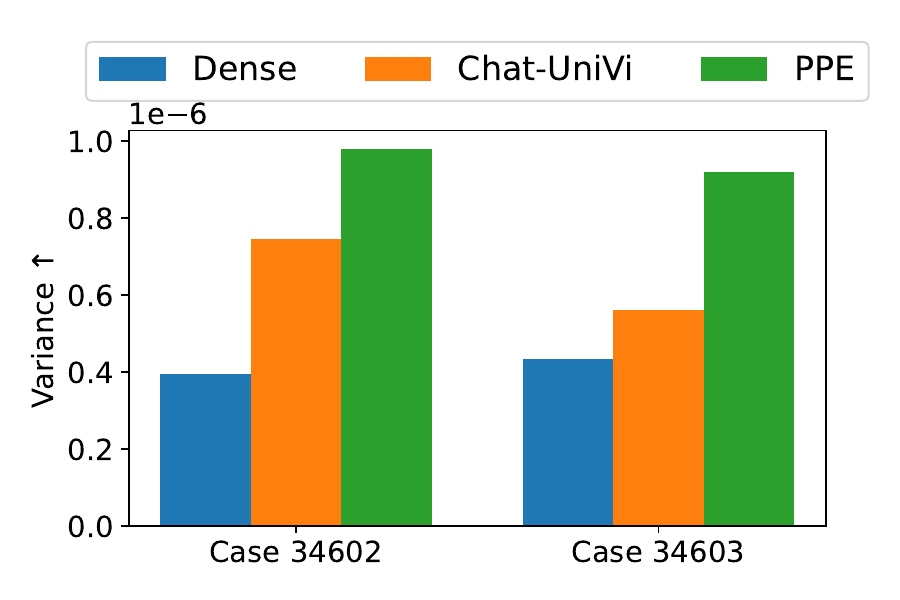}
    \caption{Variance comparison ↑}
  \end{subfigure}
  \hfill
  \begin{subfigure}{0.24\textwidth}
    \centering
    \includegraphics[width=\linewidth]{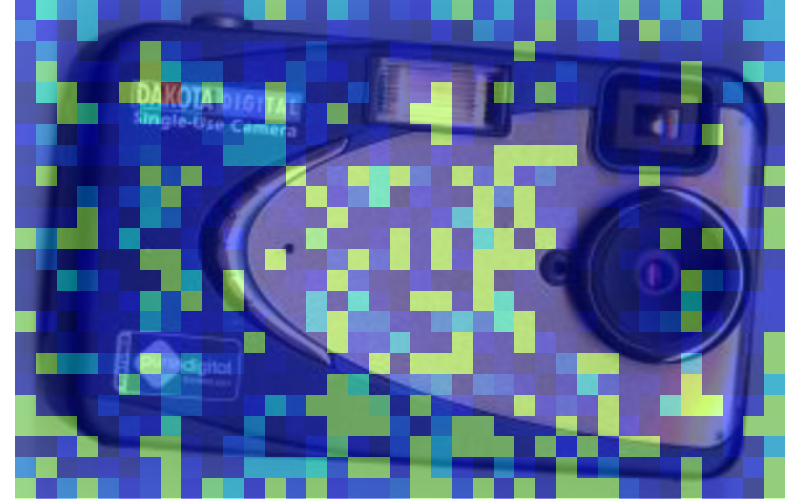}
    \caption{Chat-UniVi: "d" \ding{55}}
  \end{subfigure}
  \hfill
  \begin{subfigure}{0.24\textwidth}
    \centering
    \includegraphics[width=\linewidth]{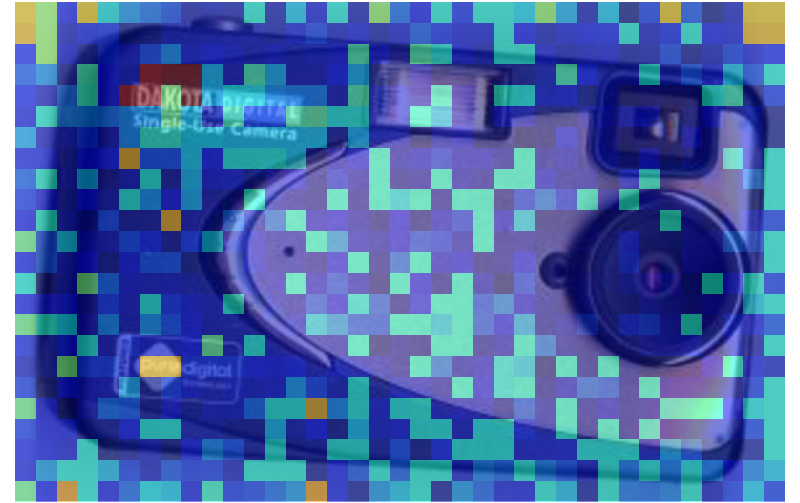}
    \caption{PPE:"dakota digital" \ding{51}}
  \end{subfigure}

  \caption{\textbf{Attention statistics and visualizations} of samples from TextVQA. 
  (a--b) Quantitative comparison of entropy and variance. 
  (c--d) Qualitative attention score visualizations of case 34602.}
  \label{fig:attn_stats}
  \vspace{-0.6em}
\end{figure*}

We randomly selected two samples and evaluated text–visual attention at the final LLM layer under a 55\% token compression ratio. As shown in Figure~\ref{fig:attn_stats}(a--b), both Chat-UniVi and PPE reduce entropy compared to the dense baseline, while PPE achieves similar entropy but consistently higher variance, indicating sharper and more confident grounding under compression. This is further illustrated in Figure~\ref{fig:attn_stats}(c--d): for case 34602 (question: \emph{"What is the brand of this camera?"}), both models attend to the correct region, but Chat-UniVi’s focus is narrowly confined due to positional information loss and answers incorrectly, whereas PPE preserves coverage across the text, recovering the full brand name. More samples and analyses are provided in \textit{Appendix~\ref{append_attn_vis}}.
\vspace{-0.6em}

\subsubsection{Integration with Other Clustering-Based Methods}

PPE is compatible with other clustering-based methods. When integrated into the training-free PACT~\cite{dhouib2025pactpruningclusteringbasedtoken} and ToMe~\cite{tome}, experiments on Qwen2-VL-7B-Instruct~\cite{wang2024qwen2vlenhancingvisionlanguagemodels} show consistent improvements over the baselines (Table~\ref{tab:other}). This compatibility arises because PPE preserves the correspondence between tokens and their RoPE components: cluster averaging keeps embeddings close, and associating merged tokens with all original RoPE indices maintains positional information. When training is allowed, PPE further improves RoPE allocation, yielding more robust joint token-position representations and enhancing performance (e.g., Chat-UniVi + PPE, Table~\ref{tab:main_table}).

To further demonstrate PPE's generality across different compression frameworks, we conduct extensive experiments on VisionZip~\cite{visionzip}, which combines pruning and merging and supports both training-free and SFT settings. We simply integrate PPE into VisionZip on Qwen2.5-VL-3B-Instruct following the official instructions ($dominant\_ratio$ and $contextual\_ratio$ are set $0.40$ and $0.05$ respectively to achieve 55\% reduction ratio), randomly sampling 1/10 of our curated dataset for the SFT setting and only fine-tuning the projector for one epoch. The results are shown in Table~\ref{tab:more_baseline}. In both settings, PPE delivers improvements in most cases, especially on layout- and OCR-sensitive benchmarks, demonstrating that PPE retains more spatial information under the same reduction ratio. Moreover, when training is allowed, PPE achieves larger gains. Our method is decoupled from model size., extensive experiments are provided in \textit{Appendix~\ref{append_model_size}}.

\begin{table}[t]
\centering
\caption{Extensive comparisons under both training-free and SFT using Qwen2.5-VL-3B-Instruct.}
\small
\begin{tabular}{@{}ccccccc@{}}
\toprule
\textbf{Setting} & \textbf{Benchmarks} & \textbf{Dense} & \textbf{Chat-UniVi} & \textbf{Chat-UniVi+PPE} & \textbf{VisionZip} & \textbf{VisionZip+PPE} \\
\cmidrule(lr){1-1} \cmidrule(lr){2-2} \cmidrule(lr){3-3} \cmidrule(lr){4-5} \cmidrule(lr){6-7}

\multirow{6}{*}{\rotatebox{90}{\textit{\textbf{Training-Free}}}} 
& \textbf{MMBench} (EN) & 79.10 & 81.50 & \textbf{82.28}~(+0.78) & \underline{\textbf{83.48}} & 83.18~(-0.30) \\
& \textbf{MMBench} (CN) & 78.10 & 80.06 & \textbf{81.43}~(+1.37) & \underline{\textbf{81.75}} & 81.64~(-0.11) \\
& \textbf{TextVQA} & 79.30 & 37.60 & \textbf{73.78}~(+36.18) & 79.00 & \underline{\textbf{79.62}}~(+0.62) \\
& \textbf{DocVQA} & \underline{93.90} & 19.58 & \textbf{66.16}~(+46.58) & 83.98 & \textbf{85.63}~(+1.65) \\
& \textbf{OCRBench} & \underline{797} & 307 & \textbf{598}~(+291) & 713 & \textbf{725}~(+12) \\
& \textbf{ChartQA} & \underline{84.00} & 18.72 & \textbf{67.08}~(+48.52) & 79.72 & \textbf{80.72}~(+1.00) \\

\midrule

\multirow{6}{*}{\rotatebox{90}{\textit{\textbf{SFT}}}} 
& \textbf{MMBench} (EN) & \underline{85.89} & \textbf{84.92} & 84.73~(-0.19) & \textbf{83.99} & 83.78~(-0.21) \\
& \textbf{MMBench} (CN) & \underline{86.07} & 83.71 & \textbf{84.87}~(+1.16) & 82.12 & \textbf{82.63}~(+0.15) \\
& \textbf{TextVQA} & 79.50 & 57.66 & \textbf{77.14}~(+19.48) & 80.02 & \underline{\textbf{82.06}}~(+2.04) \\
& \textbf{DocVQA} & 89.44 & 52.48 & \textbf{76.79}~(+24.31) & 84.84 & \underline{\textbf{90.52}}~(+5.68) \\
& \textbf{OCRBench} & 761 & 535 & \textbf{691}~(+156) & 711 & \underline{\textbf{780}}~(+69) \\
& \textbf{ChartQA} & 79.96 & 49.60 & \textbf{74.52}~(+24.92) & 80.20 & \underline{\textbf{82.36}}(+2.16) \\

\cmidrule(lr){1-7}
& \textbf{Reduction Ratio} & 0\% & 55\% & 55\% & 55\% & 55\% \\
\bottomrule
\end{tabular}
\label{tab:more_baseline}
\end{table}

\subsubsection{SFT or Training-Free}

PPE is a plug-and-play, parameter-free method. Whether SFT is required depends on the underlying compression method, not PPE itself: 1) Chat-UniVi is a merging-based method that requires SFT, and performs bad in training-free setting especially on layout-sensitive benchmarks. 2) PACT and ToMe are fully training-free, and PPE is directly applied to the official pre-trained models without any additional training. 3) VisionZip supports both training-free and SFT usage, and PPE improves performance in both settings without modifying their token selection or aggregation mechanisms. This supports our key claim that PPE can enhance existing token compression pipelines in a plug-and-play manner and can further improve RoPE allocation when training is allowed.

\subsubsection{Analysis of Inference Efficiency}

PPE introduces no additional parameters and incurs negligible computational overhead. Since it operates by redistributing existing RoPE dimensions, the computational and memory savings primarily result from token reduction itself. To ensure fairness and reproducibility, we reused the official PACT code to report runtime and memory usage, as presented in Table~\ref{tab:other}. With same reduction ratios, PPE does not need extra parameter and introduce very few inference time demonstrating its efficiency.

\begin{table}[t]
\centering
\caption{Ablation study on performance and retained ID ratio with varying $K$.}
\resizebox{\columnwidth}{!}{%
\begin{tabular}{cccccccccc}
\toprule
\textbf{$K$} & \begin{tabular}[c]{@{}c@{}}\textbf{VideoMME}\\ (w/o subs)\end{tabular} & \begin{tabular}[c]{@{}c@{}}\textbf{VideoMME}\\ (w subs)\end{tabular} & \begin{tabular}[c]{@{}c@{}}\textbf{NeXT-QA}\\ (MC)\end{tabular} & \begin{tabular}[c]{@{}c@{}}\textbf{NeXT-QA}\\ (OE)\end{tabular} & \begin{tabular}[c]{@{}c@{}}\textbf{SEED-Bench}\\ \textbf{-Video}\end{tabular} & \textbf{MVBench} & \textbf{\textit{Average}} & \begin{tabular}[c]{@{}c@{}}\textbf{Reduction}\\ \textbf{Ratio}\end{tabular} & \begin{tabular}[c]{@{}c@{}}\textbf{IDs}\\ \textbf{Retained}\end{tabular} \\
\cmidrule(lr){1-1}\cmidrule(lr){2-7}\cmidrule(lr){8-8}\cmidrule(lr){9-9}\cmidrule(lr){10-10}
\textbf{1} & 57.74 & 58.04 & 77.88 & 28.77 & 55.07 & \textbf{68.08} & \textit{57.97} & 55\% & 45\% \\
\textbf{8} & \textbf{58.70} & \textbf{59.07} & \textbf{78.42} & \textbf{32.61} & 55.98 & 67.38 & \textit{\textbf{58.69}} & 55\% & 77\% \\
\textbf{24} & 58.19 & 58.56 & 78.02 & 31.64 & 55.52 & 67.73 & \textit{58.28} & 55\% & 84\% \\
\bottomrule
\end{tabular}%
}
\label{tab:k_value}
\vspace{-0.6em}
\end{table}

\subsubsection{Effect of $K$ on Performance and Retained ID Ratio}
\label{k_value}

The hyperparameter $K$ determines how many token ID positions are retained after merging, thus affecting spatiotemporal preservation. As shown in Table~\ref{tab:k_value} and Figure~\ref{fig:idea}(b), $K=1$ causes severe degradation due to insufficient positional information, while $K=24$ introduces redundancy and slightly underperforms $K=8$. The choice of $K=8$ achieves the best trade-off, aligning with the GCD of the 3D M-RoPE dimensions [16, 24, 24] to preserve essential signals without over-retention, and is adopted as our default. Moreover, Table~\ref{tab:k_value} shows that larger $K$ values yield higher valid ID retention which refers to the ratio of distinct positional embeddings preserved after compression; for $K>1$, the retention rate exceeds the visual token retention ratio (45\%), confirming that PPE captures positional information beyond token counts. Further visualization analyses for different choices of $K$ are presented in \textit{Appendix~\ref{append_vis_k}}.

\section{Conclusion}

In this work, we propose \textbf{P}ositional \textbf{P}reservation \textbf{E}mbedding (PPE), a simple yet effective operator for retaining spatiotemporal positional information during visual token compression in MLLMs. PPE encodes fine-grained spatial and temporal cues into each compressed token and is parameter-free, plug-and-play, and compatible with existing pipelines without architectural changes. Experiments on MMBench, TextVQA, and VideoMME show consistent $2\%\sim5\%$ gains, validating its effectiveness and generality. PPE also supports cascade clustering for progressive compression, offering a flexible trade-off between efficiency and performance. Our results underscore the importance of positional information in improving MLLM efficiency.

\bibliography{iclr2026_conference}
\bibliographystyle{iclr2026_conference}

\newpage
\appendix
\section{Appendix}
\label{append}

\subsection{Analysis of Different Reduction Ratios}
\label{append_ratio}
We conduct experiments under different reduction ratios, as shown in Table~\ref{tab:ratio}. Overall, we observe that moderate compression tends to preserve the most useful positional information while eliminating redundancy. In particular, a reduction ratio of 55\% consistently yields the best results across most benchmarks. When the reduction is too aggressive (e.g., 90\%), the performance drops significantly due to the loss of critical visual content. On the other hand, smaller reductions do not fully exploit the potential for compression and efficiency. Based on these observations, we adopt a default clustering ratio of 0.45 (i.e., 55\% reduction) in all our experiments, balancing efficiency and effectiveness.

\begin{table}[h!]
\centering
\caption{Ablation study comparing different reduction ratios.}
\resizebox{\columnwidth}{!}{%
\begin{tabular}{ccccccccc}
\toprule
\textbf{Reduction Ratio} & \begin{tabular}[c]{@{}c@{}}\textbf{VideoMME}\\ (w/o subs)\end{tabular} & \begin{tabular}[c]{@{}c@{}}\textbf{VideoMME}\\ (w subs)\end{tabular} & \begin{tabular}[c]{@{}c@{}}\textbf{NeXT-QA}\\ (MC)\end{tabular} & \begin{tabular}[c]{@{}c@{}}\textbf{NeXT-QA}\\ (OE)\end{tabular} & \begin{tabular}[c]{@{}c@{}}\textbf{SEED-Bench}\\ \textbf{-Video}\end{tabular} & \textbf{MVBench} & \textbf{\textit{Average}} \\
\cmidrule(lr){1-1}\cmidrule(lr){2-7}\cmidrule(lr){8-8}
\textbf{25\%} & 58.41 & 58.22 & 78.53 & 29.37 & \textbf{56.77} & \textbf{68.03} & \textit{58.22} \\
\textbf{40\%} & 57.30 & 57.52 & 78.15 & 30.99 & 56.59 & 67.40 & \textit{57.99} \\
\textbf{55\%} & \textbf{58.70} & \textbf{59.07} & 78.42 & \textbf{32.61} & 55.98 & 67.38 & \textit{\textbf{58.69}} \\
\textbf{70\%} & 57.74 & 57.78 & \textbf{78.61} & 32.17 & 55.90 & 67.38 & \textit{58.26} &\\
\textbf{90\%} & 57.41 & 57.70 & 77.98 & 31.06 & 55.36 & 66.65 & \textit{57.69} \\ 
\bottomrule
\end{tabular}%
}
\label{tab:ratio}
\end{table}

\subsection{Analysis of Aggregation Stages}
\label{append_stages}

We conduct experiments to compare two aggregation strategies: a three-stage setting (with spatial clustering ratios 0.25/0.5/0.5) and a single-stage setting (with a clustering ratio of 0.45). As shown in Table~\ref{tab:stages}, the single-stage approach consistently outperforms the three-stage counterpart under comparable reduction ratios. Based on this observation, we adopt single-stage spatial clustering with a ratio of 0.45 as the default configuration throughout our experiments.

\begin{table}[h!]
\centering
\caption{Ablation study comparing different aggregation stages.}
\resizebox{\columnwidth}{!}{%
\begin{tabular}{ccccccccc}
\toprule
\begin{tabular}[c]{@{}c@{}}\textbf{Aggregation}\\ \textbf{Stages} \end{tabular} & \begin{tabular}[c]{@{}c@{}}\textbf{VideoMME}\\ (w/o subs)\end{tabular} & \begin{tabular}[c]{@{}c@{}}\textbf{VideoMME}\\ (w subs)\end{tabular} & \begin{tabular}[c]{@{}c@{}}\textbf{NeXT-QA}\\ (MC)\end{tabular} & \begin{tabular}[c]{@{}c@{}}\textbf{NeXT-QA}\\ (OE)\end{tabular} & \begin{tabular}[c]{@{}c@{}}\textbf{SEED-Bench}\\ \textbf{-Video} \end{tabular} & \textbf{MVBench} & \textbf{\textit{Average}} & \begin{tabular}[c]{@{}c@{}}\textbf{Reduction}\\ \textbf{Ratio} \end{tabular} \\
\cmidrule(lr){1-1}\cmidrule(lr){2-7}\cmidrule(lr){8-8}\cmidrule(lr){9-9}
\textbf{Three-Stage} & 57.30 & 57.04 & 78.14 & 32.33 & \textbf{56.06} & 66.95 & \textit{57.97} & 57\% \\
\textbf{Single-Stage} & \textbf{58.70} & \textbf{59.07} & \textbf{78.42} & \textbf{32.61} & 55.98 & \textbf{67.38} & \textit{\textbf{58.69}} & 55\% \\
\bottomrule
\end{tabular}%
}
\label{tab:stages}
\vspace{-0.8em}
\end{table}

\subsection{Analysis of Backbone Generalization} 
\label{append_backbone}
To evaluate the generalizability of PPE across backbones, we additionally test it on the LLaVA-OV-0.5B model. As shown in Table~\ref{tab:llava_result}, the Dense baseline reaches 44.8\%, while PPE achieves 44.74\% with 96\% token reduction, showing strong generalization. PPE also outperforms Chat-UniVi (43.07\%), especially in preserving spatiotemporal information.

\begin{table}[h!]
\centering
\caption{Ablation study on backbone generalization, conducted on the VideoMME (w/o subs) benchmark using the LLaVA-OV-0.5B (SI) model.}
\footnotesize
\begin{tabular}{lccccc}
\toprule
\textbf{Method} 
& \textbf{Short} & \textbf{Medium} & \textbf{Long} & \textbf{\textit{Average}} 
& \textbf{Reduction Ratio} \\
\cmidrule(lr){1-1} \cmidrule(lr){2-4} \cmidrule(lr){5-5} \cmidrule(lr){6-6}
\textbf{Dense}                     & 54.70 & \underline{43.20} & \underline{36.20} & \underline{\textit{44.80}} & 0\%  \\
\textbf{Chat-UniVi} & 53.00 & 40.40 & \textbf{35.80} & \textit{43.07} & 96\% \\
\textbf{PPE (Ours)} & \textbf{\underline{56.40}}~(+3.40) & \textbf{42.10}~(+7.70) & 35.70~(-0.10) & \textbf{\textit{44.74}}~(+1.67) & 96\% \\
\bottomrule
\end{tabular}
\label{tab:llava_result}
\end{table}

\subsection{Analysis of Model Size} 
\label{append_model_size}

Our method is decoupled from model size. To further demonstrate that PPE scales with model size, we additionally evaluate PPE on Qwen2.5-VL-7B-Instruct, results are shown in Table~\ref{tab:model_size_7b}, showing that PPE provides improvements regardless of model capacity. For the SFT setting, we trained it on only 1/5 of the data used for the 3B model due to time and resource constraints.

We also note that in the training-free setting of VisionZip, PPE shows a slight performance drop on ChartQA for the 7B model, while improving the 3B model. This may be because the stronger 7B baseline already leverages fine-grained local patterns for such task, and without SFT, the additional positional details introduced by PPE can slightly perturb the attention distribution. After SFT, PPE consistently brings positive gains. For instance, the slight drop on VisionZip+PPE for ChartQA becomes an improvement, indicating that PPE benefits more when SFT is applied.

\begin{table}[t]
\centering
\caption{Ablation study on model size using Qwen2.5-VL-7B-Instruct.}
\small
\begin{tabular}{@{}ccccccc@{}}
\toprule
\textbf{Setting} & \textbf{Benchmarks} & \textbf{Dense} & \textbf{Chat-UniVi} & \textbf{Chat-UniVi+PPE} & \textbf{VisionZip} & \textbf{VisionZip+PPE} \\
\cmidrule(lr){1-1} \cmidrule(lr){2-2} \cmidrule(lr){3-3} \cmidrule(lr){4-5} \cmidrule(lr){6-7}

\multirow{6}{*}{\rotatebox{90}{\textit{\textbf{Training-Free}}}} 
& \textbf{MMBench} (EN) & 83.50 & 83.23 & \textbf{83.58}~(+0.35) & \underline{\textbf{85.24}} & 85.19~(-0.05) \\
& \textbf{MMBench} (CN) & 83.40 & 80.18 & \textbf{82.35}~(+2.17) & 83.62 & \underline{\textbf{83.92}}~(+0.30) \\
& \textbf{TextVQA} & 84.90 & 35.82 & \textbf{63.44}~(+27.42) & 84.64 & \underline{\textbf{85.02}}~(+0.40) \\
& \textbf{DocVQA} & \underline{95.70} & 27.06 & \textbf{66.42}~(+39.36) & 87.30 & \textbf{88.52}~(+1.22) \\
& \textbf{OCRBench} & \underline{864} & 479 & \textbf{577}~(+98) & 780 & \textbf{794}~(+14) \\
& \textbf{ChartQA} & \underline{\textbf{87.30}} & 19.92 & \textbf{46.72}~(+29.90) & \textbf{60.56} & 59.88~(-0.68) \\

\midrule

\multirow{6}{*}{\rotatebox{90}{\textit{\textbf{SFT}}}} 
& \textbf{MMBench} (EN) & \textbf{86.90} & 86.23 & \textbf{86.26}~(+0.03) & 86.14 & \underline{\textbf{87.24}}~(+0.16) \\
& \textbf{MMBench} (CN) & \underline{\textbf{85.35}} & 84.25 & \textbf{84.85}~(+0.60) & 85.12 & \textbf{85.15}~(0.03) \\
& \textbf{TextVQA} & 87.20 & 54.92 & \textbf{82.46}~(+27.54) & 87.08 & \underline{\textbf{87.24}}~(+0.16) \\
& \textbf{DocVQA} & 92.97 & 50.01 & \textbf{85.84}~(+35.83) & 93.16 & \underline{\textbf{93.57}}~(+0.41) \\
& \textbf{OCRBench} & 826 & 584 & \textbf{764~(+180)} & \underline{\textbf{828}} & \underline{\textbf{828}}~(+0) \\
& \textbf{ChartQA} & \underline{\textbf{86.32}} & 43.96 & \textbf{78.88}~(+34.92) & 85.40 & \textbf{86.08}~(+0.68) \\

\cmidrule(lr){1-7}
& \textbf{Reduction Ratio} & 0\% & 55\% & 55\% & 55\% & 55\% \\
\bottomrule
\end{tabular}
\label{tab:model_size_7b}
\end{table}

\subsection{Analysis of Task Generalization}
\label{append_task_generalization}

PPE is inherently task-agnostic and can be seamlessly integrated into any clustering-based token merging method, regardless of downstream task. While our main experiments focus on MLLM QA tasks—largely due to their popularity and the convenience of evaluation—we emphasize that PPE is not limited to QA scenarios. 

To further demonstrate its generalization capability, we conducted additional experiments on the image captioning task using Flickr30k~\cite{flicker30k} benchmark. We reused the same models from Table~\ref{tab:main_table} and directly evaluated their zero-shot captioning performance without any further fine-tuning. The results are in Table~\ref{tab:flicker}. 

\begin{table}[h!]
\centering
\resizebox{\linewidth}{!}{
\begin{tabular}{lcccccccc}
\toprule
                   & Bleu\_1 & Bleu\_2 & Bleu\_3 & Bleu\_4 & METEOR & ROUGE\_L & SPICE & CIDEr \\
\midrule
\textbf{Dense}        & 0.311 & $\underline{0.188}$ & $\underline{0.120}$ & $\underline{0.079}$ & $\underline{0.147}$ & $\underline{0.322}$ & $\underline{0.219}$ & $\underline{0.722}$ \\
\textbf{+ Chat-UniVi} & 0.280 & 0.167               & 0.104               & 0.068               & 0.137               & 0.311               & \textbf{0.214}      & 0.649               \\
\textbf{+ PPE}        & \textbf{\underline{0.313}}~(+0.033) & \textbf{0.185}~(+0.018) & \textbf{0.117}~(+0.013) & \textbf{0.076}~(+0.008) & \textbf{0.144}~(+0.007) & \textbf{0.312}~(+0.001) & 0.211~(-0.003) & \textbf{0.690}~(+0.041) \\
\bottomrule
\end{tabular}
}
\caption{Ablation study on task generalization beyond QA.}
\label{tab:flicker}
\end{table}

From the table above, as expected, Dense achieves the best overall performance. Chat-UniVi, without PPE, suffers a noticeable performance drop. For instance, CIDEr decreases from 0.722 to 0.649, reflecting reduced content relevance. BLEU scores also decline, suggesting degraded coherence due to spatial misalignment caused by compression. SPICE shows a slight decrease (0.219 → 0.214), indicating minor semantic loss. With PPE, performance is largely restored across most metrics: BLEU-1/4 improve from 0.280/0.068 to 0.313/0.076, surpassing Dense in BLEU-1, and CIDEr rises from 0.649 to 0.690, narrowing the gap with Dense. Additionally, qualitative cases further illustrate PPE's effectiveness, as presented in Appendix~\ref{append_flicker30}.

Both quantitative and qualitative results confirm PPE's ability to recover structural and positional cues often lost during naive token merging. While it does not fully close the gap with Dense, PPE consistently mitigates performance degradation.

\subsection{Qualitative Cases on Image Captioning}
\label{append_flicker30}
\begin{figure}[htbp]
  \centering
  \begin{subfigure}{0.45\textwidth}
    \centering
    \includegraphics[width=\linewidth]{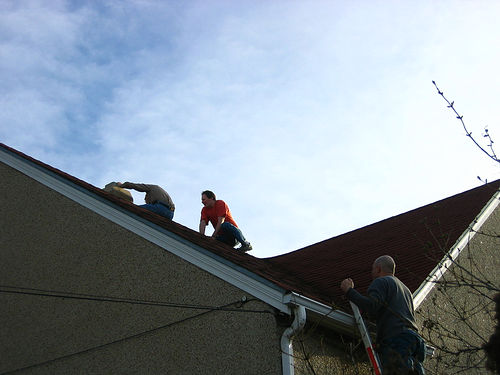}
    \caption{Case 1: 10287332, roof scene}
  \end{subfigure}
  \hfill
  \begin{subfigure}{0.45\textwidth}
    \centering
    \includegraphics[width=\linewidth]{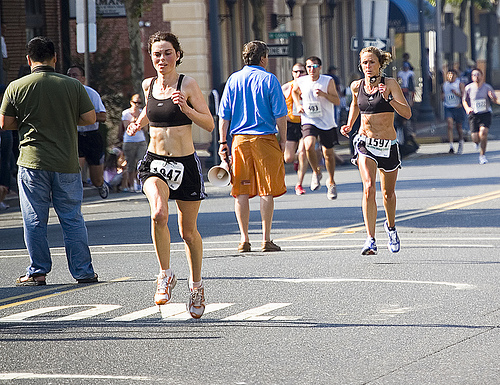}
    \caption{Case 2: 900144365, marathon scene}
  \end{subfigure}
  \caption{\textbf{Qualitative examples for the image captioning task}, from Flickr30k benchmark.}
  \label{fig:appendix_flicker30}
\end{figure}

We randomly select two representative cases from Flickr30k, as shown in Figure~\ref{fig:appendix_flicker30}. 

\textbf{Case 1 (10287332.jpg).} The task prompt is: ``Provide a one-sentence caption for the provided image. Do not provide any explanation.'' The ground truth caption is: ``Two men sitting on the roof of a house while another one stands on a ladder.'' The Dense model predicts: ``Three men work on a roof with a blue sky in the background.'' Under compression, Chat-UniVi simplifies it to: ``Three men work on a roof,'' omitting background and clothing cues. In contrast, PPE preserves richer details: ``Three men working on a roof with one wearing an orange shirt,'' showing stronger spatial recall under compression.  

\textbf{Case 2 (900144365.jpg).} The ground truth caption is: ``Marathon runners are racing on a city street, with other people standing around.'' The Dense model predicts: ``A woman with number 1397 is running in a race.'' Chat-UniVi predicts: ``A woman with number 1947 on her shirt is running in a race.'' PPE outputs: ``A woman with the number 247 on her shirt is running in a race.'' All models focus primarily on the runners' numbers. However, there are multiple runners, including men and women, making the predictions less accurate. In the foreground, two women runners are the main focus of the image. The number of the front runner is only partially visible as ``27'', while the runner behind wears ``1597''. Consequently, the Dense model appears to focus on the runner behind and all models misrecognize the number.

\subsection{Attention Visualization Analysis}
\label{append_attn_vis}

Using the same models from Table~\ref{tab:main_table}, we extract the attention maps of the final LLM layer (layer\_id=35) between all text and visual tokens and project them back onto the original images as heatmaps, revealing each question's focus across image regions. Both Chat-UniVi and PPE are under a reduction ratio of 55\%. We randomly selected some cases to demonstrate our PPE can preserve valid positional information and focus more on the key region of images.

\begin{figure*}[htbp]
  \centering
  \begin{subfigure}{0.3\textwidth}
    \centering
    \includegraphics[width=\linewidth]{figures//appendix/1505.png}
    \caption{Original}
  \end{subfigure}
  \hfill
  \begin{subfigure}{0.3\textwidth}
    \centering
    \includegraphics[width=\linewidth]{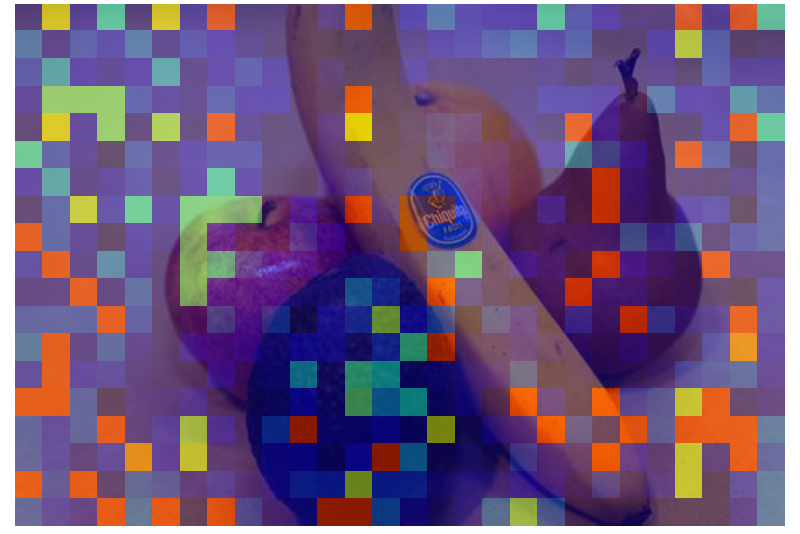}
    \caption{Chat-UniVi}
  \end{subfigure}
  \hfill
  \begin{subfigure}{0.3\textwidth}
    \centering
    \includegraphics[width=\linewidth]{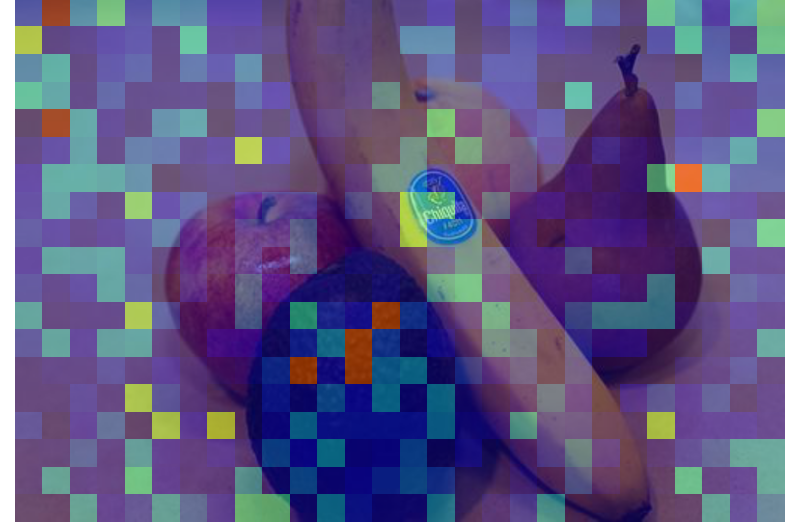}
    \caption{PPE}
  \end{subfigure}

  \begin{subfigure}{0.3\textwidth}
    \centering
    \includegraphics[width=\linewidth]{figures//appendix/37890-af67237a4a504a29.jpg}
    \caption{Original}
  \end{subfigure}
  \hfill
  \begin{subfigure}{0.3\textwidth}
    \centering
    \includegraphics[width=\linewidth]{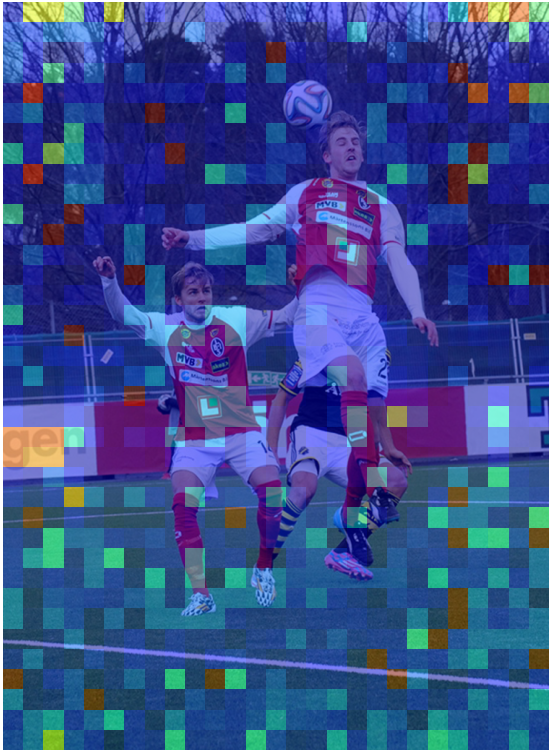}
    \caption{Chat-UniVi}
  \end{subfigure}
  \hfill
  \begin{subfigure}{0.3\textwidth}
    \centering
    \includegraphics[width=\linewidth]{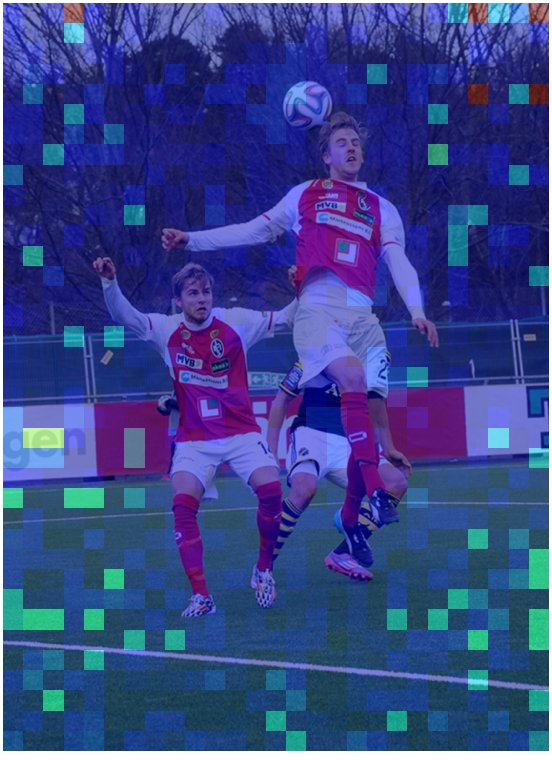}
    \caption{PPE}
  \end{subfigure}

  \begin{subfigure}{0.3\textwidth}
    \centering
    \includegraphics[width=\linewidth]{figures//appendix/37376-23183bf2db16d88c.jpg}
    \caption{Original}
  \end{subfigure}
  \hfill
  \begin{subfigure}{0.3\textwidth}
    \centering
    \includegraphics[width=\linewidth]{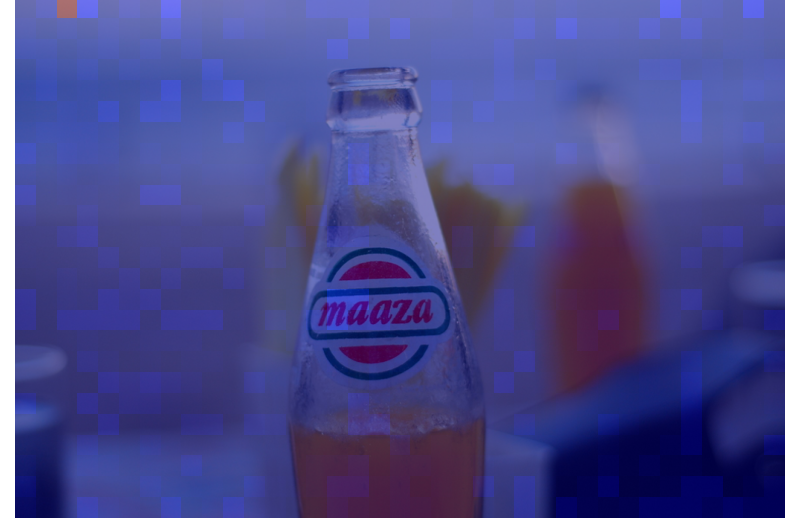}
    \caption{Chat-UniVi}
  \end{subfigure}
  \hfill
  \begin{subfigure}{0.3\textwidth}
    \centering
    \includegraphics[width=\linewidth]{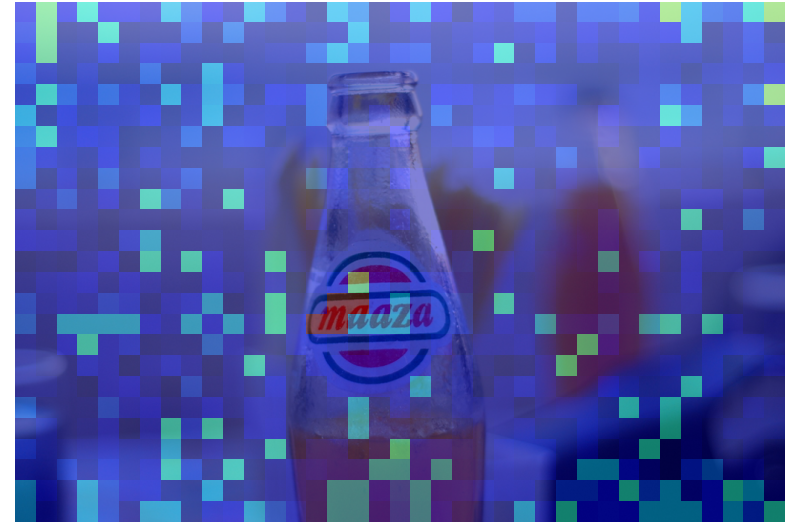}
    \caption{PPE}
  \end{subfigure}

  \caption{\textbf{Qualitative comparison of attention visualizations.} 
  Each row corresponds to a different sample: (a, d, g) original input, (b, e, h) Chat-UniVi outputs, and (c, f, i) PPE outputs.}
  \label{fig:attn_vis}
\end{figure*}

\textbf{Case 1. MMBench (EN), (a--c) in Figure~\ref{fig:attn_vis}.}  
Official QA index: \texttt{1505}. Question: \emph{"How many types of fruits are there in the image? Please give the answer directly. A. 3 B. 2 C. 5 D. 4 E. None."}  
The image contains five fruit types: a central banana, surrounded by an apple, avocado, pear, and a partially visible orange in the background. The correct answer is \emph{C}. PPE predicts correctly, while Chat-UniVi outputs \emph{D}.  

Attention heatmaps show that Chat-UniVi partly focuses on the background and fails to capture the orange, likely leading to the incorrect prediction. In contrast, PPE successfully attends to all fruit regions, including the occluded orange, demonstrating better spatial alignment under token compression and improved grounding for object counting in complex scenes.  

\textbf{Case 2. TextVQA, (d--f) in Figure~\ref{append_attn_vis}.}  
Official image name: \texttt{af67237a4a504a29.jpg}, question ID: 37890. Question: \emph{"What number is visible on the shorts in front?"}  
The image depicts three football players on a grassy field. The player in the foreground, jumping mid-air, wears shorts with the number "2". Behind him, another player shows a partially visible "14", of which only the "1" is clear. The correct answer is \emph{2}. Both Chat-UniVi and PPE incorrectly output \emph{1}.  

Heatmaps indicate that models attend to the "2" on the foreground player, the partial "1" from the second player, and even the text "gen" on an advertising board. PPE places more emphasis on the "2", while Chat-UniVi is more distracted by background text. Although both fail in this case, the richer attention of PPE may sometimes introduce ambiguity, but also reflects broader contextual awareness.  

\textbf{Case 3. TextVQA, (g--i) in Figure~\ref{append_attn_vis}.}  
Official image name: \texttt{23183bf2db16d88c.jpg}, question ID: 37376. Question: \emph{"What kind of drink is this?"}  
The image shows an orange beverage bottle labeled with the brand \emph{maaza}. The ground-truth answer is \emph{maaza}. Both the Dense model and Chat-UniVi predict correctly, while PPE outputs \emph{"orange juice"}.  

Visualizations reveal that PPE’s attention covers both the brand logo and the broader bottle region. Although its answer deviates from the ground truth, it remains semantically reasonable, suggesting that PPE’s broader positional retention can capture additional contextual cues beyond the exact label.

\subsection{Limitations and Failure Cases}
\label{append_failure_cases}

\begin{figure}[htbp]
  \centering
  \begin{subfigure}{0.5\textwidth}
    \centering
    \includegraphics[width=\linewidth]{figures//appendix/textvqa_failure_23183bf2db16d88c.jpg}
    \caption{Case 1}
  \end{subfigure}
  \hfill
  \begin{subfigure}{0.3\textwidth}
    \centering
    \includegraphics[width=\linewidth]{figures//appendix/textvqa_failure_0b9a494c78985373.jpg}
    \caption{Case 2}
  \end{subfigure}
  \caption{\textbf{Representative failure cases of PPE}, from TextVQA benchmark.}
  \label{fig:appendix_failure}
\end{figure}

As shown in Tables~\ref{tab:main_table} and~\ref{tab:ratio}, PPE underperforms Dense in certain cases, highlighting the trade-off between compression and fidelity. We visualize representative failure cases to illustrate PPE's limitations.

Using the same models from Table~\ref{tab:main_table} (Qwen2.5-VL-3B-Instruct), we extract attention maps from the final LLM layer (layer\_id=35) between text and visual tokens and visualize them as heatmaps. Under identical compression ratios (55\% and 90\%), we analyze two failure examples from the TextVQA benchmark:

\textbf{Case 1 (55\% compression).} Official image: 23183bf2db16d88c.jpg; Question: ``What kind of drink is this?'' (question ID: 37376). The image shows an orange beverage labeled ``maaza''. Ground truth: ``maaza''. Dense and Chat-UniVi predict correctly, while PPE outputs ``orange juice''. Heatmaps indicate that PPE attends to both the logo and the bottle, producing a semantically relevant but less precise answer. This suggests that PPE’s positional retention captures broader context, though at the cost of exactness.

\textbf{Case 2 (90\% compression).} Official image: 0b9a494c78985373.jpg; Question: ``What word is written on the collar of the jacket?'' (question ID: 36877). The image contains multiple words: ``kutxa'' (collar), ``menabi'' (chest), ``gipuzkoa'', and partial ``deba''. Ground truth: ``kutxa''. Dense predicts correctly; Chat-UniVi outputs ``menabi''; PPE outputs ``deb?''. Heatmaps show that both Chat-UniVi and PPE attend to larger central regions, missing the smaller collar text. This demonstrates a limitation of PPE under aggressive compression, where maintaining spatial precision becomes challenging.

\subsection{PPE Gains across Different Image Resolutions}
\label{append_ppe_gains_resolutions}

\begin{figure}[t]
    \centering
    \includegraphics[width=0.7\linewidth]{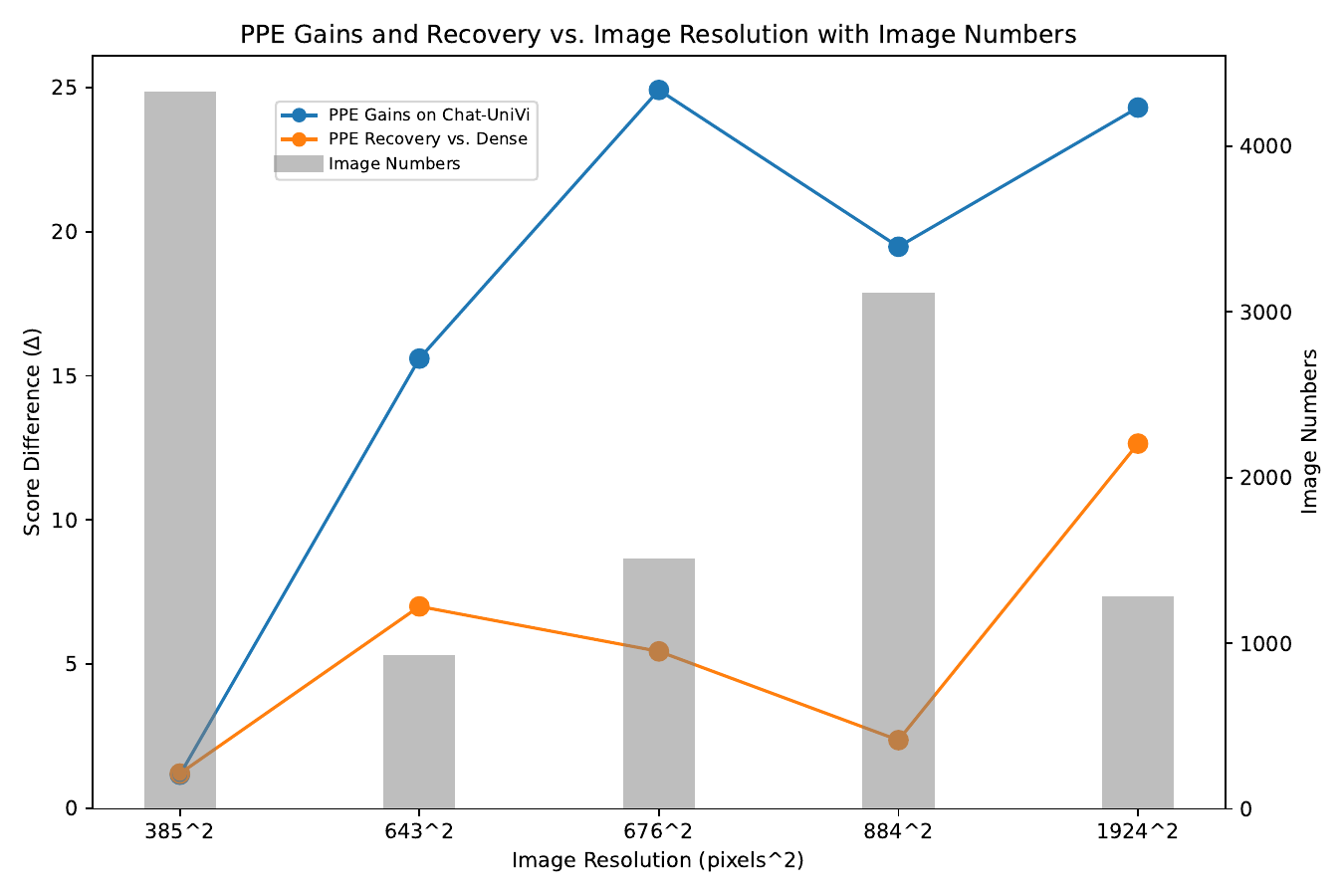}
    \caption{PPE obtains consistent improvements across different resolutions.}
    \label{fig:ppe_resolution}
\end{figure}

To evaluate the effect of PPE under varying image resolutions, we calculated the average resolution of TextVQA, MMBench, DocVQA, OCRBench, and ChartQA, based on the resolution statistics of their samples. As illustrated in Figure~\ref{fig:ppe_resolution}, the horizontal axis denotes the average image resolution for each dataset, while the left vertical axis presents the performance difference ($\Delta$). This metric reflects both the improvements introduced by PPE over Chat-UniVi and its ability to recover performance relative to the dense model. For OCRBench, whose raw score range is not from 0 to 100, scores are normalized by a factor of 10 to ensure consistent visualization. The right vertical axis represents the number of image samples in each dataset.

From the figure, we observe that PPE consistently provides notable performance gains across a wide spectrum of image resolutions: from $385^2$ to $1924^2$. These results demonstrate that PPE imposes no limitations on input resolution and remains robust and effective across images of different scales.

\subsection{Visualization of the Effect of $K$}
\label{append_vis_k}

To better understand how different values of $K$ influence attention behavior, we conduct a visual analysis and support it with quantitative indicators such as entropy and variance. Following the analytical framework introduced in Section~\ref{sec_ppe_insight} and Figure~\ref{fig:attn_stats}, we compare different choices of $K$ directly on the images.

\begin{figure*}[t]
  \vspace{-0.6em}
  \centering

  \begin{subfigure}{0.45\textwidth}
    \centering
    \includegraphics[width=\linewidth]{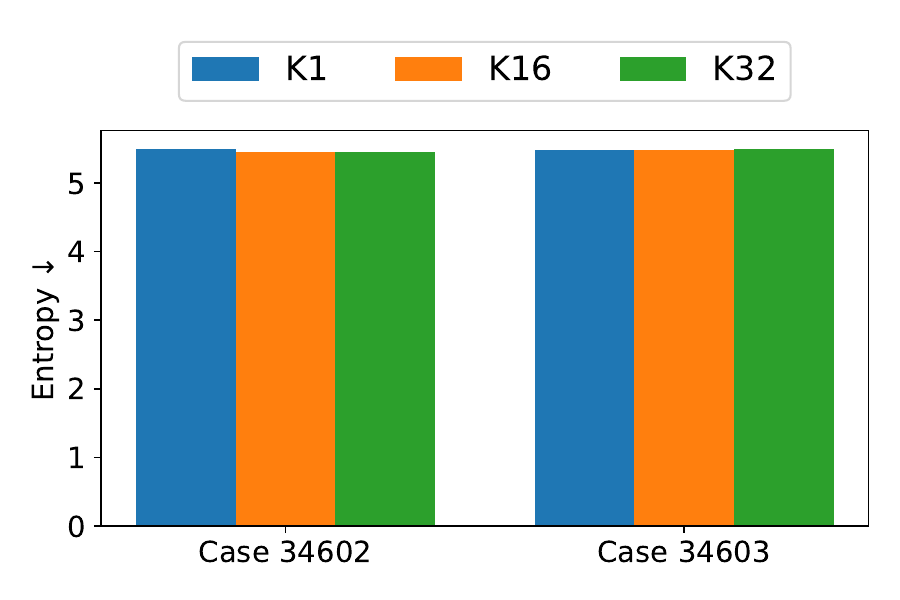}
    \caption{Entropy comparison ↓}
  \end{subfigure}
  \hfill
  \begin{subfigure}{0.45\textwidth}
    \centering
    \includegraphics[width=\linewidth]{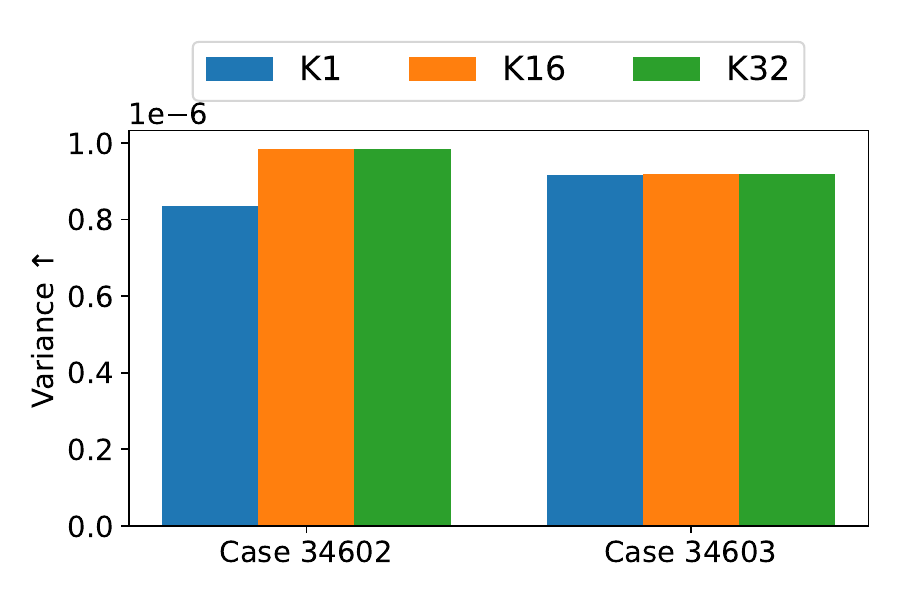}
    \caption{Variance comparison ↑}
  \end{subfigure}

  \vspace{0.7em}

  \begin{subfigure}{0.30\textwidth}
    \centering
    \includegraphics[width=\linewidth]{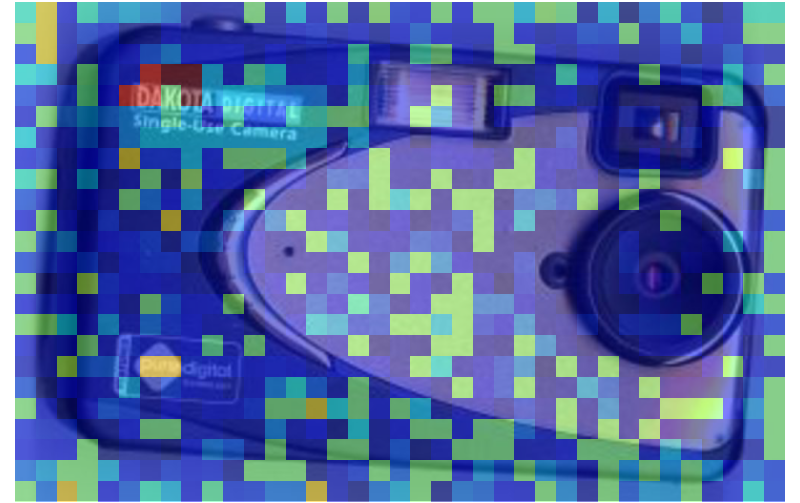}
    \caption{PPE (K=1)}
  \end{subfigure}
  \hfill
  \begin{subfigure}{0.30\textwidth}
    \centering
    \includegraphics[width=\linewidth]{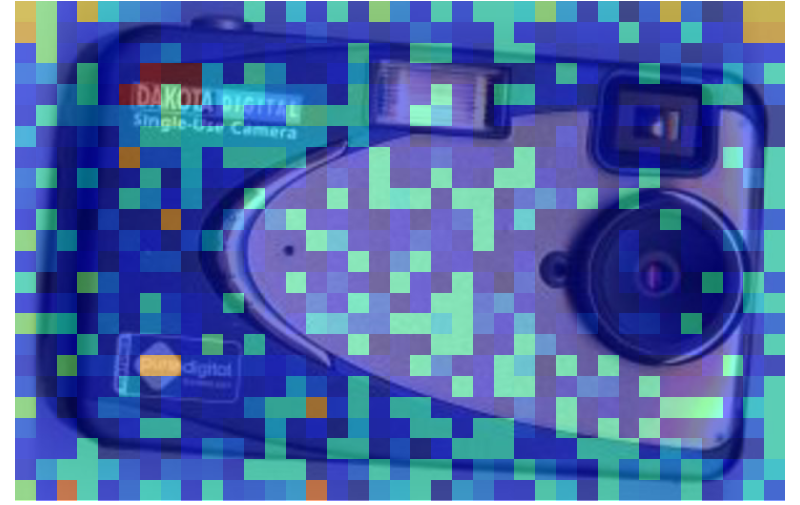}
    \caption{PPE (K=16)}
  \end{subfigure}
  \hfill
  \begin{subfigure}{0.30\textwidth}
    \centering
    \includegraphics[width=\linewidth]{figures//ppe_34602_attn.png}
    \caption{PPE (K=32)}
  \end{subfigure}

  \caption{\textbf{Attention statistics and visualizations} of samples from TextVQA.
  (a--b) Quantitative comparison of entropy and variance.
  (c--e) Qualitative attention score visualizations of case 34602.}
  \label{fig:attn_stats_k}
  \vspace{-0.6em}
\end{figure*}

As shown in Figure~\ref{fig:attn_stats_k}, when $K=1$, entropy is slightly higher and variance is lower, indicating that the model’s attention becomes more dispersed compared with $K=16$ or $K=32$. The visualizations further reveal that at $K=1$, the model mainly focuses on the first half of the brand text (with brighter activations in red). Meanwhile, tokens on the camera backplate also become more activated, showing that attention is distracted by background regions and fails to capture the full brand string, unlike the cases with $K=16$ or $K=32$. For images, we set $K=32$ based on the GCD of M-RoPE configuration $[32,32]$, which maximizes the preservation of positional information per merged token. The same rationale applies to video settings, $K=8$ of $[16,24,24]$.

M-RoPE splits embedding dimensions into multiple periodic components (temporal, vertical, horizontal), assigning each axis contiguous frequency bands. Existing analysis (e.g., Figure 3 and 4 in Reference~\cite{huang2025revisiting}) shows that uneven allocation can lead to some axes occupying only high- or low-frequency channels, impairing long-range and spatial modeling. Choosing K based on the GCD ensures that each merged token preserves multiple positional anchors evenly across the frequency spectrum, preventing collapse into only high- or low-frequency bands. In short, $K$ is grounded in M-RoPE’s frequency structure, providing theoretical justification for PPE’s ability to preserve positional information after token merging.

In short, refer to the analysis of positional encoding in VLMs~\cite{huang2025revisiting}, choosing an appropriate $K$ ensures that positional indices are distributed evenly across frequency bands; otherwise, some indices may collapse into only high- or low-frequency ranges, degrading positional fidelity.

\end{document}